\documentclass[10pt,twocolumn,letterpaper]{article}
\usepackage[pagenumbers]{iccv}              
\usepackage{graphicx}
\usepackage{booktabs}
\usepackage{amsmath}
\usepackage{amssymb}
\usepackage{multirow}
\usepackage{array}
\usepackage{makecell}
\usepackage{bm}
\usepackage{cuted}
\usepackage{capt-of}
\usepackage{pifont}

\newcommand*{\affaddr}[1]{#1} 
\newcommand*{\affmark}[1][*]{\textsuperscript{#1}}
\newcommand*{\email}[1]{\texttt{#1}}

\definecolor{iccvblue}{rgb}{0.21,0.49,0.74}
\usepackage[pagebackref,breaklinks,colorlinks,allcolors=iccvblue]{hyperref}

\title{PAN-Crafter: Learning Modality-Consistent Alignment for PAN-Sharpening}

\author{
Jeonghyeok Do\affmark[1]\quad Sungpyo Kim\affmark[1]\quad Geunhyuk Youk\affmark[1]\quad Jaehyup Lee\affmark[2]\footnotemark[2]\quad Munchurl Kim\affmark[1]\footnotemark[2]\\[0.5em]
\affaddr{\affmark[1]KAIST} \quad
\affaddr{\affmark[2]Kyungpook National University}\\
\small{\email{\{ehwjdgur0913, ksp04204, rmsgurkjg, mkimee\}@kaist.ac.kr}} \quad
\small{\email{jaehyuplee@knu.ac.kr}}\\
\small{\url{https://kaist-viclab.github.io/PAN-Crafter_site}}
}

\begin{document}
\maketitle

\begin{strip}\centering
\vspace{-1.5cm}
    \includegraphics[width=1.0\linewidth]{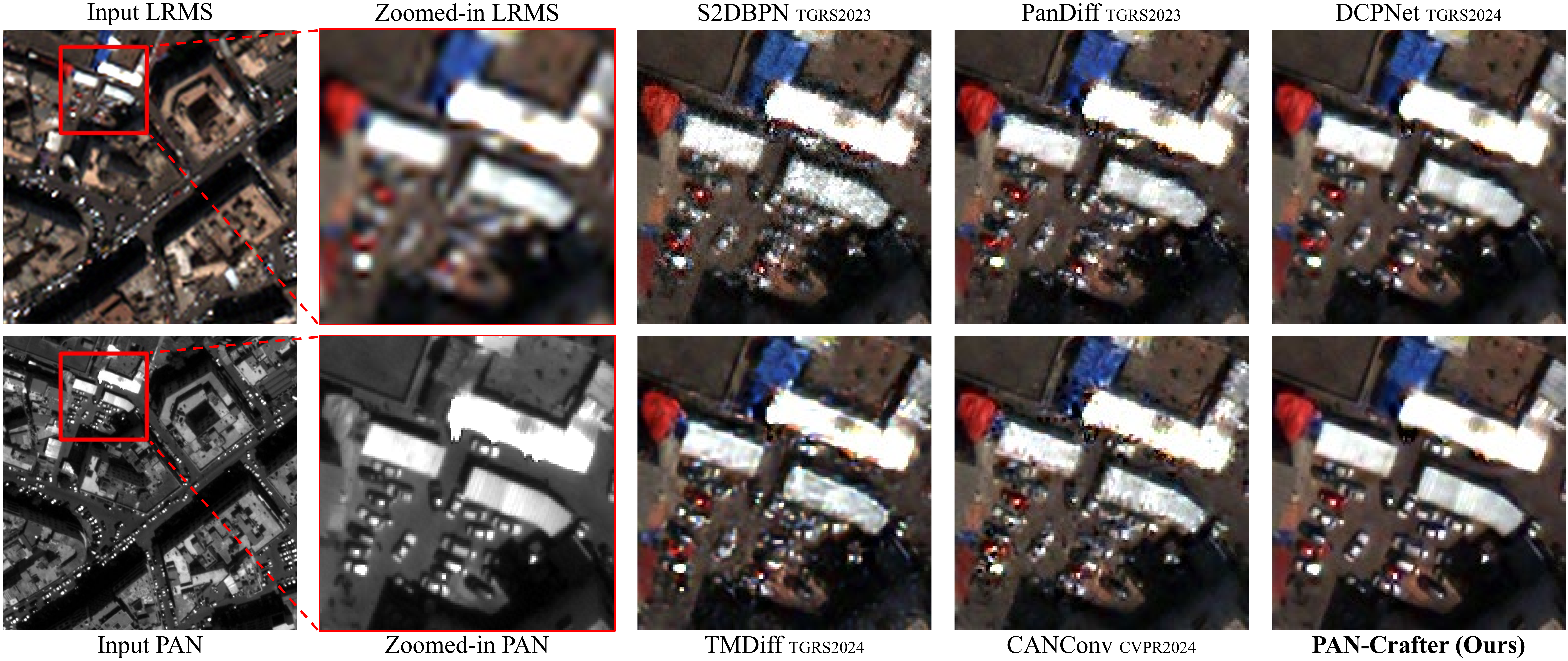}
    \vspace{-0.7cm}
    \captionof{figure}{Comparison of PAN-sharpening (PS) results on the WV3 dataset at full-resolution for very recent methods and our PAN-Crafter. The \textit{top-left} image shows the input low-resolution multi-spectral (LRMS) image with a zoomed-in region for better visualization. The \textit{bottom-left} image represents the corresponding panchromatic (PAN) image. The remaining results shows the zoomed-in PAN-sharped patches that are their corresponding restored high-resolution multi-spectral (HRMS) ones. Our proposed PAN-Crafter produces pan-sharpened images with minimal artifacts, especially near buildings and cars, while other approaches often yield blurred or distorted results.}
    \label{fig:first}
\end{strip}

{
  \renewcommand{\thefootnote}%
    {\fnsymbol{footnote}}
  \footnotetext[2]{Co-corresponding authors (equal advising).}
}
\vspace{-5mm}

\begin{abstract}

PAN-sharpening aims to fuse high-resolution panchromatic (PAN) images with low-resolution multi-spectral (MS) images to generate high-resolution multi-spectral (HRMS) outputs. However, cross-modality misalignment---caused by sensor placement, acquisition timing, and resolution disparity---induces a fundamental challenge. Conventional deep learning methods assume perfect pixel-wise alignment and rely on per-pixel reconstruction losses, leading to spectral distortion, double edges, and blurring when misalignment is present. To address this, we propose PAN-Crafter, a modality-consistent alignment framework that explicitly mitigates the misalignment gap between PAN and MS modalities. At its core, Modality-Adaptive Reconstruction (MARs) enables a single network to jointly reconstruct HRMS and PAN images, leveraging PAN’s high-frequency details as auxiliary self-supervision. Additionally, we introduce Cross-Modality Alignment-Aware Attention (CM3A), a novel mechanism that bidirectionally aligns MS texture to PAN structure and vice versa, enabling adaptive feature refinement across modalities. Extensive experiments on multiple benchmark datasets demonstrate that our PAN-Crafter outperforms the most recent state-of-the-art method in all metrics, even with 50.11$\times$ faster inference time and 0.63$\times$ the memory size. Furthermore, it demonstrates strong generalization performance on unseen satellite datasets, showing its robustness across different conditions.

\end{abstract}
\vspace{-0.4cm}    
\section{Introduction}
\label{sec:intro}

Remote sensing imagery is crucial for a wide range of applications, including environmental monitoring, defense intelligence, and urban planning \cite{yang2013role, xu2023ai, li2020review, yuan2020deep, neyns2022mapping, wellmann2020remote, do2024c}. Many of these tasks require high-resolution images that preserve both fine spatial details and rich spectral information. However, due to inherent limitations in sensor technologies, a single imaging system cannot simultaneously achieve high spatial resolution and high spectral fidelity. To overcome this constraint, modern Earth observation satellites employ dual-sensor systems, consisting of a high-resolution panchromatic (PAN) sensor and a low-resolution multi-spectral (LRMS) sensor.

\begin{figure}[tbp]
    \centering
    \includegraphics[width=1.0\columnwidth]{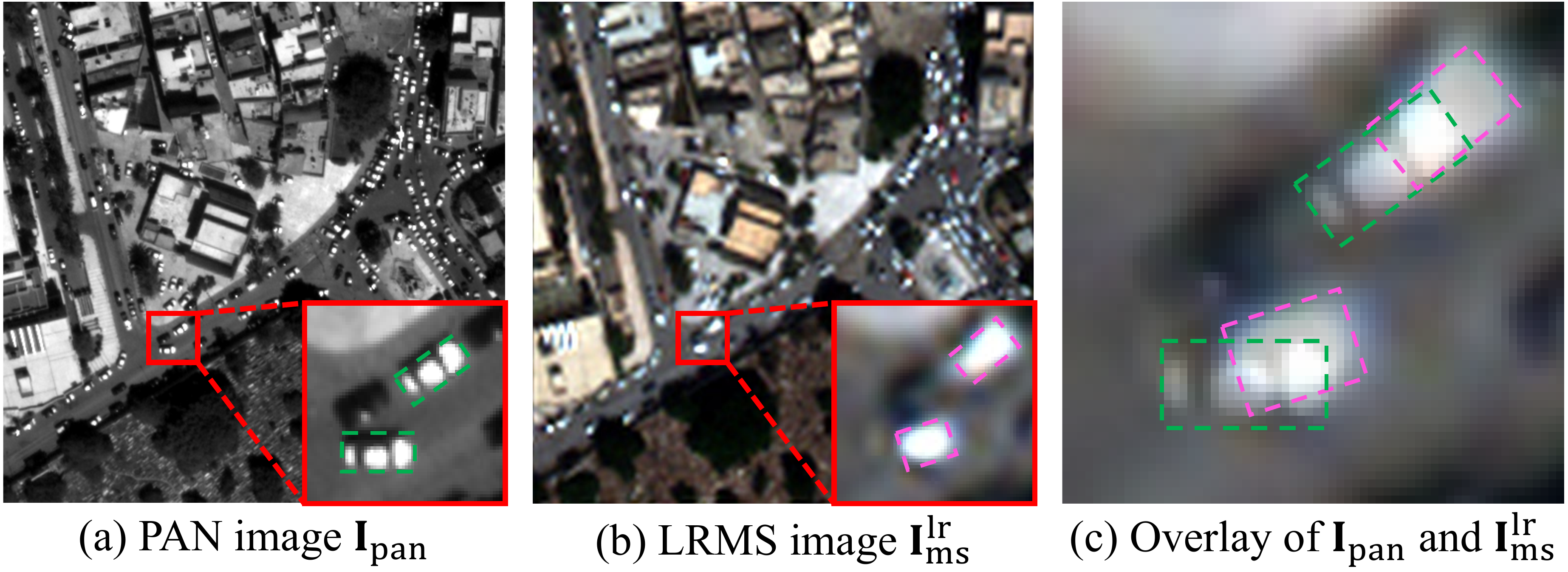}
    \vspace{-0.6cm}
    \caption{Example of PAN, LRMS, and their overlayed visualization. (a) High-resolution PAN image $\mathbf{I}_\text{pan}$, (b) Low-resolution multi-spectral image $\mathbf{I}_\text{ms}^\text{lr}$, (c) Overlay of PAN and LRMS images to highlight differences. The red insets provide zoomed-in views for better visualization.}
    \vspace{-0.5cm}
    \label{fig:motiv}
\end{figure}

PAN-sharpening \cite{deng2022machine, meng2020large, vivone2020new, loncan2015hyperspectral, thomas2008synthesis, zhu2023probability} aims to fuse high-resolution PAN images with low-resolution MS images to produce high-resolution multi-spectral (HRMS) outputs. The goal is to retain the spectral fidelity of MS images while preserving the spatial details of PAN images. However, a fundamental challenge in this process is cross-modality misalignment, which arises from differences in sensor placement, acquisition timing, and resolution disparity. As illustrated in Fig.~\ref{fig:motiv}, PAN images typically have four times ($4H\times4W$) the spatial resolution ($H\times W$) of MS images, necessitating up-sampling on the MS images before fusion. However, this up-sampling step introduces interpolation artifacts and spatial shifts, further amplifying alignment discrepancies. Most existing PAN-sharpening methods \cite{wang2021ssconv, wu2021dynamic, zhu2023probability, peng2024fusionmamba, meng2023pandiff, zhang2024dcpnet, duan2024content} assume perfect pixel-wise alignment and rely on per-pixel reconstruction losses, such as $\ell_1$ and $\ell_2$, leading to spectral distortion, double edges, and blurring when misalignment is present. To address these issues, several approaches \cite{lee2021sipsa, duan2024content, jin2022lagconv} integrate spatial-adaptive convolutional layers to mitigate misalignment. Despite these efforts, existing approaches lack the capability to dynamically adapt to varying levels of misalignment across different datasets. Fixed-scale alignment mechanisms fail to capture complex spatial shifts \cite{lee2021sipsa}, while self-similarity-based feature aggregation does not explicitly correct geometric discrepancies between PAN and MS images \cite{duan2024content, jin2022lagconv}. Effective PAN-sharpening requires a solution that not only aligns textures and structures across modalities but also ensures modality-consistent feature refinement at multiple spatial scales.

To overcome these limitations, we propose PAN-Crafter, a modality-consistent alignment framework designed to handle cross-modality misalignment during the fusion process. Unlike existing methods that assume strict pixel-wise alignment, PAN-Crafter enables robust learning from misaligned PAN-MS pairs by jointly reconstructing HRMS and PAN images, ensuring structural consistency through feature alignment. Our key innovation is Modality-Adaptive Reconstruction (MARs), which allows a single network to dynamically generate both HRMS and PAN images based on a modality selection mechanism, MARs mode. Given LRMS and PAN as inputs, the network reconstructs HRMS in MS mode and PAN in PAN mode, using PAN’s sharpness as an auxiliary self-supervision signal to enhance spatial fidelity. Furthermore, we introduce Cross-Modality Alignment-Aware Attention (CM3A), a novel mechanism that explicitly (i) aligns MS image textures to PAN image structures during HRMS reconstructions and (ii) matches PAN image structures to MS image textures during PAN back-reconstructions. This bidirectional interaction enables adaptive compensation for misalignment while preserving both spatial and spectral integrity. Our main contributions are as follows:

\begin{itemize}
    \item We propose Modality-Adaptive Reconstruction (MARs), a unified reconstruction framework that enables robust learning from misaligned PAN-MS image pairs by dynamically generating both HRMS and PAN images;
    \item We introduce Cross-Modality Alignment-Aware Attention (CM3A), a novel alignment mechanism that adaptively refines textures and structures between PAN and MS images, improving spatial-spectral consistency;
    \item We achieve state-of-the-art (SOTA) performance across multiple benchmark datasets and show strong robustness on unseen satellite datasets, demonstrating the effectiveness of our PAN-Crafter in handling real-world cross-modality misalignment.

\end{itemize}

\section{Related Work}
\label{sec:related}

\subsection{PAN-Sharpening}

\noindent\textbf{Traditional approaches.} Traditional PAN-sharpening (PS) methods are broadly classified into component substitution (CS) \cite{choi2010new, shensa1992discrete, carper1990use, rahmani2010adaptive}, multi-resolution analysis (MRA) \cite{aiazzi2006mtf, otazu2005introduction, aiazzi2002context, shah2008efficient}, and variational optimization (VO) \cite{ballester2006variational, wu2021vo+, fu2019variational, cao2022proximal}.

\noindent\textbf{Deep learning-based approaches.} 
Recent advancements in PAN-sharpening have been driven by deep learning-based methods \cite{peng2024fusionmamba, hou2024linearly, hou2023bidomain, bandara2022hypertransformer, zhu2023probability, duan2024content, zhang2024dcpnet}, primarily leveraging Convolutional Neural Networks (CNNs) \cite{o2015introduction}. CNN-based models \cite{masi2016pansharpening, lee2021sipsa, yang2017pannet, yuan2018multiscale, masi2016pansharpening, wu2021dynamic, jin2022lagconv, zhang2023spatial, yang2022memory, duan2024content, zhang2024dcpnet} are effective in capturing local spatial-spectral dependencies while maintaining relatively low computational complexity. For instance, S2DBPN \cite{zhang2023spatial} introduces a spatial-spectral back-projection framework to iteratively refine high-resolution outputs, while DCPNet \cite{zhang2024dcpnet} formulates a dual-task learning strategy that integrates PAN-sharpening with super-resolution. More recently, diffusion-based models \cite{meng2023pandiff, xing2024empower, kim2024u, zhong2024ssdiff} have emerged, leveraging iterative denoising processes to refine reconstructed images. While diffusion models \cite{ho2020denoising} improve generation quality, they suffer from excessive computational costs, limiting their practical deployment in real-world applications.

\noindent\textbf{Cross-modality misalignment handling.} Most existing PAN-sharpening methods assume perfect pixel-wise alignment and optimize reconstruction using per-pixel losses. However, real-world PAN-MS pairs often exhibit spatial misalignment, leading to spectral distortion and double-edge artifacts. To address this, several works have explored spatial-adaptive feature alignment strategies. SIPSA \cite{lee2021sipsa} explicitly identifies misalignment as a critical challenge in PAN-sharpening and introduces a spatially-adaptive module, but its fixed-scale alignment mechanism limits flexibility when handling diverse misalignment patterns. LAGConv \cite{jin2022lagconv} and CANConv \cite{duan2024content} adopt non-local spatial-adaptive convolution modules that enhance feature consistency through self-similarity. However, these methods primarily focus on semantic feature aggregation rather than explicit geometric alignment, making them suboptimal for correcting cross-modality shifts.

\subsection{Spatial-Adaptive Operation}

PAN-sharpening datasets are generally pre-aligned, but local misalignment persists due to sensor inconsistencies, parallax effects, and resolution differences. Addressing these spatial shifts requires spatial-adaptive operations that dynamically adjust processing based on input features, enabling local refinement while maintaining structural consistency. Several adaptive techniques have been proposed to tackle local misalignment by modulating computations according to spatial context. Pixel-adaptive convolution \cite{su2019pixel} dynamically adjusts convolutional weights based on local pixel intensity, enabling spatially adaptive filtering. Deformable convolutions \cite{dai2017deformable} learn spatially adaptive offsets, shifting receptive fields to enhance feature extraction. However, these methods primarily focus on intra-modality feature refinement and lack explicit mechanisms to enforce cross-modality alignment.

\noindent \textbf{Attention mechanisms.} Attention operations \cite{vaswani2017attention, dosovitskiy2020image, pan2023slide, Zn5r_1, Zn5r_2, Zn5r_3, Zn5r_4} are inherently non-local and spatially adaptive, enabling dynamic feature aggregation across spatial regions. Self-attention mechanisms capture long-range dependencies by computing pairwise feature relationships, while cross-attentions extend this by integrating information across different modalities. To incorporate spatial priors, positional embeddings \cite{vaswani2017attention, liu2021swin} are typically added to query and key representations, encoding relative pixel positions that guide feature interactions. However, in cross-modality settings, fixed positional embeddings fail to handle local misalignment, as real-world distortions vary across datasets and cannot be effectively modeled by static spatial priors. 

To address these limitations, we replace fixed positional embeddings with modality-aware feature priors directly derived from PAN and MS representations. Instead of encoding positional information explicitly, our approach leverages cross-modality feature interactions by embedding PAN features into queries ($\mathbf{Q}$) and keys ($\mathbf{K}$) when attending to MS, and vice versa. This design enables the PS networks to dynamically adapt spatial attention to local misalignment patterns, ensuring precise feature alignment without reliance on predefined spatial encodings.

\begin{figure*}[tbp]
    \centering
    \includegraphics[width=1.0\textwidth]{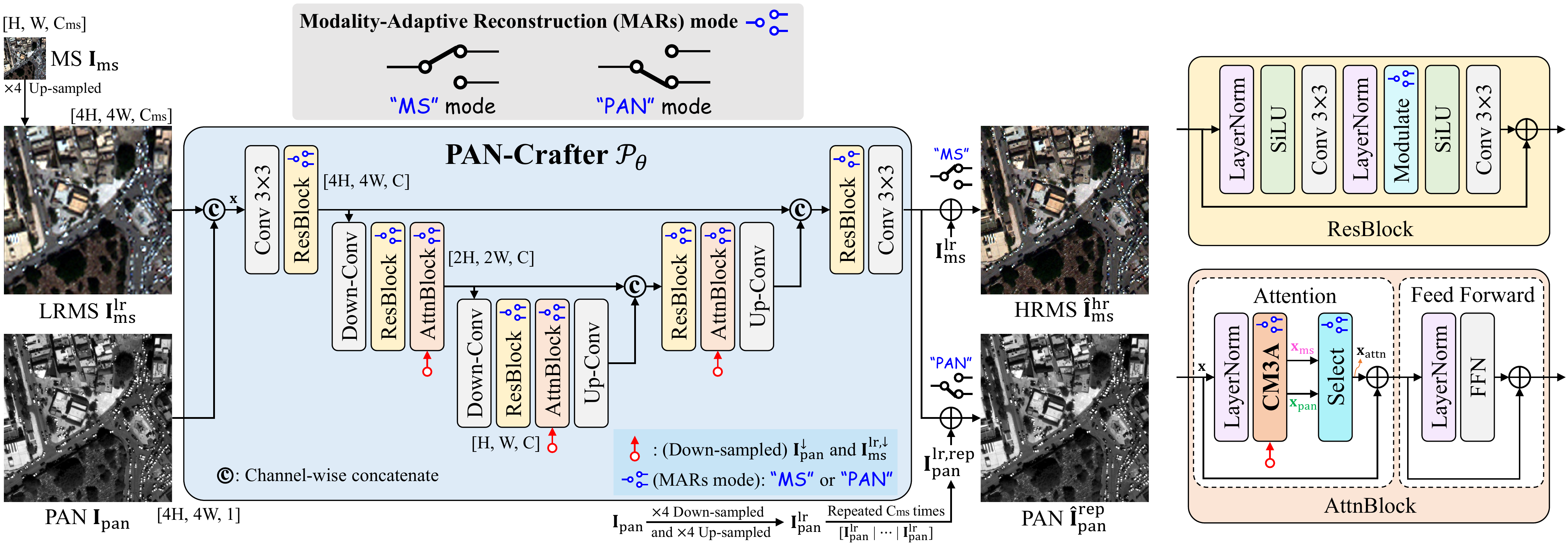}
    \vspace{-0.6cm}
    \caption{Overview of the proposed PAN-Crafter architecture. The network processes input PAN and LRMS images leveraging Modality-Adaptive Reconstruction (MARs) mode, which enables adaptive generation of HRMS and PAN outputs. By leveraging spatial structures and spectral fidelity of PAN and MS images, MARs ensures high-frequency details are preserved while minimizing spectral distortion. The architecture follows an encoder-decoder design, incorporating residual blocks and cross-modality alignment-aware attention (CM3A) at multiple scales to mitigate modality misalignment while preserving spectral and structural fidelity of MS and PAN images, respectively.}
    \label{fig:main}
    \vspace{-0.3cm}
\end{figure*}
\section{Methods}
\label{sec:method}

\subsection{Overview of the Proposed PAN-Crafter}
\label{sec:31}

Given a paired dataset \( \mathcal{D} = \{(\mathbf{I}_{\text{pan}}, \mathbf{I}_{\text{ms}}, \mathbf{I}_{\text{ms}}^{\text{hr}})\} \), where \( \mathbf{I}_{\text{pan}} \in \mathbb{R}^{4H \times 4W \times 1} \) represents a PAN image, and \( \mathbf{I}_{\text{ms}} \in \mathbb{R}^{H \times W \times C_{\text{ms}}} \) denotes an MS image with \( C_{\text{ms}} \) spectral bands. Since the MS image has a lower spatial resolution, we first up-sample it by a factor of \(4\), obtaining a LRMS \( \mathbf{I}_{\text{ms}}^{\text{lr}} \in \mathbb{R}^{4H \times 4W \times C_{\text{ms}}} \), which serves as an initial estimate of HRMS target \( \mathbf{I}_{\text{ms}}^{\text{hr}} \in \mathbb{R}^{4H \times 4W \times C_{\text{ms}}}\).

\noindent \textbf{Modality-Adaptive Reconstruction (MARs).} Our goal is to learn a PAN-Crafter network \( \mathcal{P}_{\theta} \) that synthesizes a HRMS image \( \mathbf{I}_{\text{ms}}^{\text{hr}} \) from the given PAN and LRMS inputs while explicitly handling cross-modality misalignment. To achieve this, we introduce Modality-Adaptive Reconstruction (MARs), a dynamic mechanism that enables the network to selectively reconstruct either HRMS or PAN images depending on the modality selection (MARs mode). By jointly learning to reconstruct both HRMS and PAN images within a shared network, PAN-Crafter effectively incorporates sharp spatial structures while maintaining spectral fidelity in the HRMS prediction. In \(\mathsf{MS}\) mode, the PS network $\mathcal{P}_{\theta}$ learns to align structural details from \( \mathbf{I}_{\text{pan}} \) to \( \mathbf{I}_{\text{ms}}^{\text{lr}} \) using Cross-Modality Alignment-Aware Attention (CM3A), predicting the final HRMS output $\hat{\mathbf{I}}_{\text{ms}}^{\text{hr}}$ as:

\begin{equation}
    \hat{\mathbf{I}}_{\text{ms}}^{\text{hr}} = \mathcal{P}_{\theta} \left(\mathbf{I}_{\text{pan}}, \mathbf{I}_{\text{ms}}^{\text{lr}} ; \text{mode}=\mathsf{MS}\right) + \mathbf{I}_{\text{ms}}^{\text{lr}}.
\end{equation}

\noindent Conversely, in \(\mathsf{PAN}\) mode, $\mathcal{P}_{\theta}$ predicts a multi-channel version of PAN, defined as:
\vspace{-0.1cm}

\begin{equation}
\mathbf{I}_{\text{pan}}^{\text{rep}}=\left[\mathbf{I}_{\text{pan}} \mid \cdots \mid \mathbf{I}_{\text{pan}}\right] \in \mathbb{R}^{4H \times 4W \times C_{\text{ms}}},
\end{equation}

\noindent where $[\;\cdot\mid\cdot\;]$ denotes channel-wise concatenation, ensuring consistency across spectral bands. Since $\mathbf{I}_{\text{pan}}$ is a single-channel image, but $\mathbf{I}_{\text{ms}}^{\text{lr}}$ consists of \( C_{\text{ms}} \) spectral bands, we formulate the PAN back-reconstruction as:
\vspace{-0.3cm}

\begin{equation}
    \hat{\mathbf{I}}_{\text{pan}}^{\text{rep}} = \mathcal{P}_{\theta} \left(\mathbf{I}_{\text{pan}}, \mathbf{I}_{\text{ms}}^{\text{lr}}; \text{mode}=\mathsf{PAN}\right) + \mathbf{I}_{\text{pan}}^{\text{lr}, \text{rep}},
\end{equation}

\noindent where \( \mathbf{I}_{\text{pan}}^{\text{lr}} \) is obtained by first down-sampling the original PAN image \( \mathbf{I}_{\text{pan}} \) by a factor of \( 4 \), followed by up-sampling it back to the original resolution and \(\mathbf{I}_{\text{pan}}^{\text{lr, rep}}\) represents the LR PAN image \(\mathbf{I}_{\text{pan}}^{\text{lr}}\) replicated \(C_{\text{ms}}\) times channel-wise, as illustrated in Fig.~\ref{fig:main} (bottom). This formulation aligns with the residual learning strategy in \(\mathsf{MS}\) mode, ensuring that the network in \(\mathsf{PAN}\) mode learns high-frequency refinements. At inference time, we fix the MARs mode to $\mathsf{MS}$ mode, ensuring that the network always produces \( \hat{\mathbf{I}}_{\text{ms}}^{\text{hr}} \) as the final HRMS output.

\noindent \textbf{MARs loss.} To ensure stable training, we duplicate each $(\mathbf{I}_{\text{pan}}, \mathbf{I}_{\text{ms}}, \mathbf{I}_{\text{ms}}^{\text{hr}})$ triplet along the batch dimension, processing one instance in \(\mathsf{MS}\) mode and the other in \(\mathsf{PAN}\) mode. This enables $\mathcal{P}_{\theta}$ to learn modality-specific features while maintaining consistency across both reconstruction tasks. For each mode, we apply the $\ell_1$ loss independently to the predicted HRMS image $\hat{\mathbf{I}}_{\text{ms}}^{\text{hr}}$ and the back-reconstructed multi-channel PAN image $ \hat{\mathbf{I}}_{\text{pan}}^{\text{rep}}$, ensuring spatial and spectral fidelity. Our MARs loss function is defined as:
\vspace{-0.3cm}

\begin{equation}
    \mathcal{L}_{\text{MARs}} = \left\|\hat{\mathbf{I}}_{\text{ms}}^{\text{hr}} - \mathbf{I}_{\text{ms}}^{\text{hr}}\right\|_1 + \lambda\left\|\hat{\mathbf{I}}_{\text{pan}}^{\text{rep}} - \mathbf{I}_{\text{pan}}^{\text{rep}}\right\|_1,
\end{equation}

\noindent where $\lambda$ is a weighting factor that balances the contribution of the PAN back-reconstruction loss. Since PAN images contain rich high-frequency details, tuning $\lambda$ ensures that $\mathcal{P}_{\theta}$ effectively incorporates sharp spatial structures while maintaining spectral fidelity in the MS reconstruction.

\subsection{PAN-Crafter}
\label{sec:32}

The proposed PAN-Crafter $\mathcal{P}_{\theta}$ is a U-Net-based network designed for robust cross-modality feature alignment and high-quality HRMS reconstruction. As shown in Fig.~\ref{fig:main}, the network follows an encoder-decoder architecture where each stage consists of a combination of residual blocks (ResBlocks) and cross-modality alignment blocks (AttnBlocks). To effectively handle varying levels of misalignment, we integrate Cross-Modality Alignment-Aware Attention (CM3A) at multiple scales throughout the network. Specifically, low- and mid-resolution stages incorporate both ResBlock and AttnBlock in a cascaded manner, while high-resolution stages use only ResBlock to reduce computational overhead.

\noindent\textbf{Network Architecture.} $\mathcal{P}_{\theta}$ takes as input the channel-wise concatenation of \(\mathbf{I}_{\text{pan}}\) and \(\mathbf{I}_{\text{ms}}^{\text{lr}}\). A convolutional layer ($\mathsf{Conv}$) first embeds the input into a feature representation \( \mathbf{x} \) of channel dimension \( C \). The feature \( \mathbf{x} \) then passes through multiple encoder stages that consist of down-sampling convolutional layers ($\mathsf{Down}\text{-}\mathsf{Conv}$), ResBlocks, and AttnBlocks. 
Then, \( \mathbf{x} \) is progressively decoded using up-sampling layers ($\mathsf{Up}\text{-}\mathsf{Conv}$), ResBlocks, and AttnBlocks while preserving structural details and spectral integrity.

\noindent\textbf{Residual Block (ResBlock).} The ResBlock is designed to refine modality-specific features while preserving spatial structures. As illustrated in Fig.~\ref{fig:main}, each ResBlock consists of Layer Normalization ($\mathsf{LN}$), $\mathsf{SiLU}$ activation, and $\mathsf{Conv}$ as:
\vspace{-0.7cm}

\begin{equation}
    \begin{split}
    \mathbf{x} &\leftarrow \mathsf{Conv}\left(\mathsf{SiLU}\left(\mathsf{LN}\left(\mathbf{x}\right)\right)\right),\\
    \mathbf{x} &\leftarrow \mathbf{x} + \mathsf{Conv}\left(\mathsf{SiLU}\left(\mathsf{Modulate}\left(\mathsf{LN}\left(\mathbf{x}\right); \text{mode}\right)\right)\right),
    \end{split}
\end{equation}

\noindent where $\mathsf{Modulate}$ is a feature modulation layer. We incorporate a modulation mechanism that adjusts channel-wise feature scaling based on the input modality as:
\vspace{-0.4cm}

\begin{equation}
    \begin{split}
    \mathsf{Modulate}\left(\mathbf{x}; \mathsf{MS}\right):& \;\mathbf{x} \leftarrow (1 + \bm{\gamma}_{\text{ms}}) \odot \mathbf{x} + \bm{\beta}_{\text{ms}},\\
    \mathsf{Modulate}\left(\mathbf{x}; \mathsf{PAN}\right):& \;\mathbf{x} \leftarrow (1 + \bm{\gamma}_{\text{pan}}) \odot \mathbf{x} + \bm{\beta}_{\text{pan}},\\
    \end{split}
\end{equation}

\noindent where $\bm{\gamma}_{\text{ms}}, \bm{\beta}_{\text{ms}}, \bm{\gamma}_{\text{pan}}, \bm{\beta}_{\text{pan}} \in \mathbb{R}^C$ are learnable parameters, and $\odot$ denotes channel-wise multiplication. This ensures modality-aware feature adaptation while maintaining structural consistency.

\noindent\textbf{Cross-Modality Attention Block (AttnBlock).} The AttnBlock is designed to facilitate modality-aware feature interaction while preserving structural consistency. As illustrated in Fig.~\ref{fig:main}, the block consists of two key components: an attention layer and a feed-forward layer. The attention layer employs the CM3A to dynamically align features between the PAN and MS modalities as:
\vspace{-0.2cm}

\begin{equation}
\mathbf{x}_{\text{ms}},\;\mathbf{x}_{\text{pan}} = \text{CM3A}\left(\mathsf{LN}\left(\mathbf{x}\right);\text{mode}\right).
\end{equation}

\noindent Subsequently, the selection layer ($\mathsf{Select}$) integrates complementary information as:
\vspace{-0.2cm}

\begin{equation}
    \begin{split}
    \mathbf{x}_{\text{attn}} &= \bm{\alpha}_{\text{ms}}^{1} \odot \mathbf{x}_{\text{ms}} + \bm{\alpha}_{\text{ms}}^{2} \odot \mathbf{x}_{\text{pan}} \;(\mathsf{MS} \;\text{mode}),\\
    \mathbf{x}_{\text{attn}} &= \bm{\alpha}_{\text{pan}}^{1} \odot \mathbf{x}_{\text{ms}} + \bm{\alpha}_{\text{pan}}^{2} \odot \mathbf{x}_{\text{pan}} \;(\mathsf{PAN} \;\text{mode}),
    \end{split}
\end{equation}

\noindent where $\bm{\alpha}_{\text{ms}}^{1}, \bm{\alpha}_{\text{ms}}^{2}, \bm{\alpha}_{\text{pan}}^{1}, \bm{\alpha}_{\text{pan}}^{2} \in \mathbb{R}^C$ are learnable parameters. The final attended feature is integrated via a residual connection as $\mathbf{x} \leftarrow \mathbf{x} + \mathbf{x}_{\text{attn}}$. Following the attention operation, the feed-forward network ($\mathsf{FFN}$) refines the attended features as $\mathbf{x} \leftarrow \mathbf{x} + \mathsf{FFN}\left(\mathsf{LN}\left(\mathbf{x}\right)\right)$.

\subsection{Cross-Modality Alignment-Aware Attention}
\label{sec:33}

\begin{figure}[tbp]
    \centering
    \includegraphics[width=1.0\columnwidth]{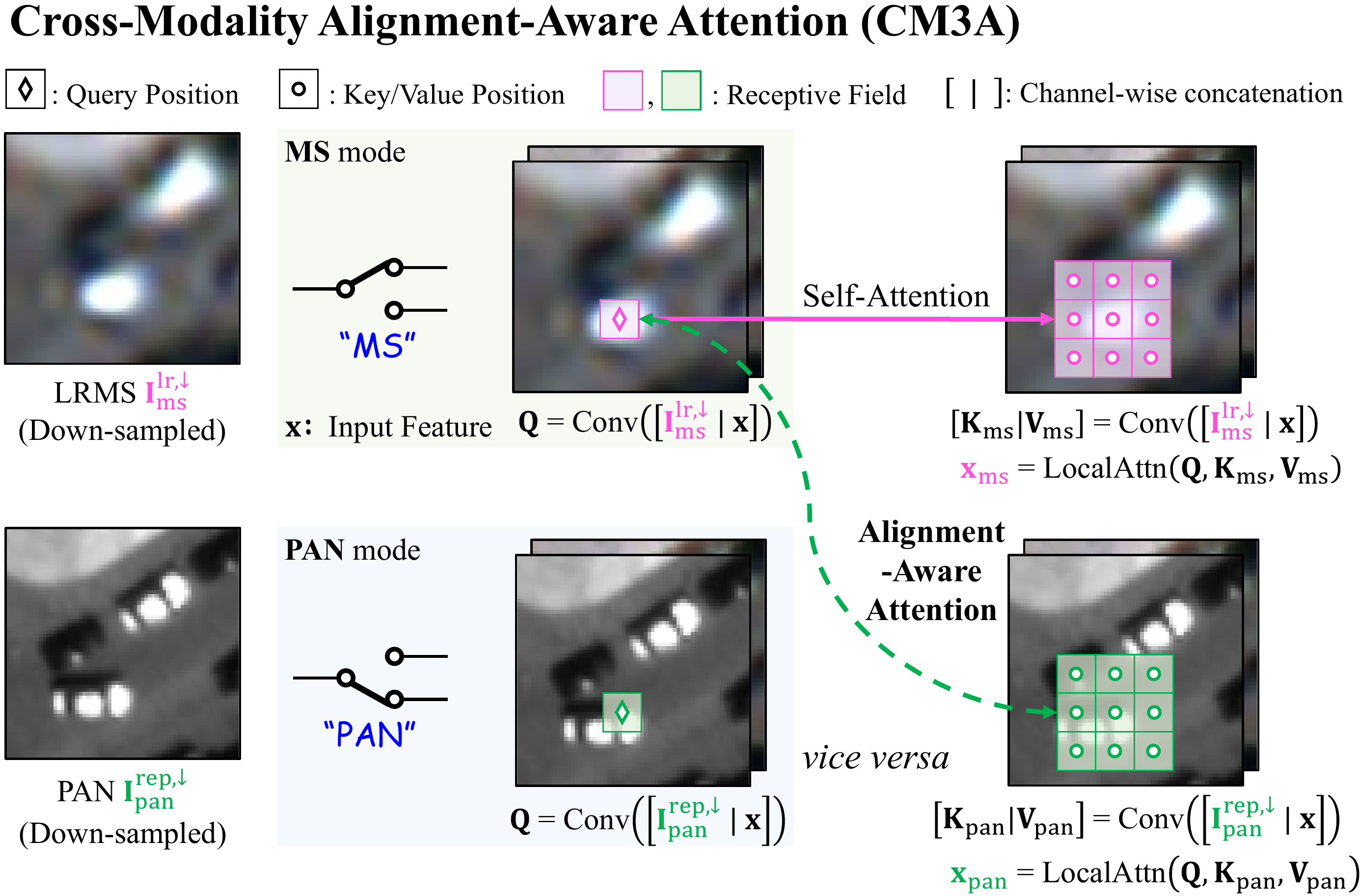}
    \vspace{-0.6cm}
    \caption{Cross-Modality Alignment-Aware Attention (CM3A) enables bidirectional alignment by transferring MS texture to PAN structure during HRMS reconstruction and PAN structure to MS texture during PAN back-reconstruction. This mechanism not only mitigates cross-modality misalignment but also ensures structural and spectral fidelity in the reconstructed images.}
    \label{fig:attn}
    \vspace{-0.4cm}
\end{figure}

\noindent \textbf{Local attention mechanism.} To effectively handle cross-modality misalignment, it is not necessary to estimate the global displacement across the entire image. Since $(\mathbf{I}_{\text{pan}}, \mathbf{I}_{\text{ms}}, \mathbf{I}_{\text{ms}}^{\text{hr}})$ triplets are generally pre-aligned to a certain degree, our CM3A adopts Pan \textit{et al.} \cite{pan2023slide} and operates within a local attention window rather than global attention. As shown in Fig.~\ref{fig:attn}, for a given query position $ (i, j) $, we compute attention scores only within a $ k \times k $ receptive field centered around the query, ensuring computational efficiency while capturing local misalignment, where $ k=2k^{\prime}+1 $ is the receptive field size. Given a query feature $ \mathbf{Q} \in \mathbb{R}^{H \times W \times C} $, and key-value pairs $ \mathbf{K}, \mathbf{V} \in \mathbb{R}^{H \times W \times C} $ , Local Attention function ($\mathsf{LocalAttn}$) \cite{pan2023slide} computes attention scores within the $ k \times k $ local receptive field.

\noindent \textbf{Misalignment-guided feature interaction.} Our CM3A dynamically aligns features based on the selected MARs mode by integrating both self-attention and alignment-aware attention mechanisms. As illustrated in Fig.~\ref{fig:attn}, the attention process differs depending on whether $\mathcal{P}_\theta$ operates in $\mathsf{MS}$ mode or $\mathsf{PAN}$ mode. In $\mathsf{MS}$ mode, $\mathcal{P}_\theta$ aims to predict $\mathbf{I}_{\text{ms}}^{\text{hr}}$, ensuring spectral fidelity while incorporating structural details from PAN images. To achieve this, self-attention is first applied to maintain consistency within the MS feature space. Specifically, the query feature $\mathbf{Q}$ is constructed by concatenating the input feature $\mathbf{x}$ with a down-sampled version of the LRMS image $\mathbf{I}_{\text{ms}}^{\text{lr},\downarrow}$, ensuring that both have the same spatial resolution:
\vspace{-0.2cm}

\begin{equation}
\mathbf{Q} = \mathsf{Conv}\left(\left[\mathbf{I}_{\text{ms}}^{\text{lr},\downarrow} \mid \mathbf{x}\right]\right).
\end{equation}

\noindent The query attends to key-value pairs derived from the same modality, enabling self-attention to refine MS-specific features. $\mathbf{K}_{\text{ms}}$ and $\mathbf{V}_{\text{ms}}$ are constructed as:

\begin{equation}
    \begin{split}
    & \left[\mathbf{K}_{\text{ms}} \mid \mathbf{V}_{\text{ms}}\right] = \mathsf{Conv}\left(\left[\mathbf{I}_{\text{ms}}^{\text{lr},\downarrow} \mid \mathbf{x}\right]\right),\\
    & \mathbf{x}_{\text{ms}} = \mathsf{LocalAttn}(\mathbf{Q}, \mathbf{K}_{\text{ms}}, \mathbf{V}_{\text{ms}}).\\
    \end{split}
\end{equation}

\begin{figure*}[tbp]
  \centering
  \includegraphics[width=1.0\textwidth]{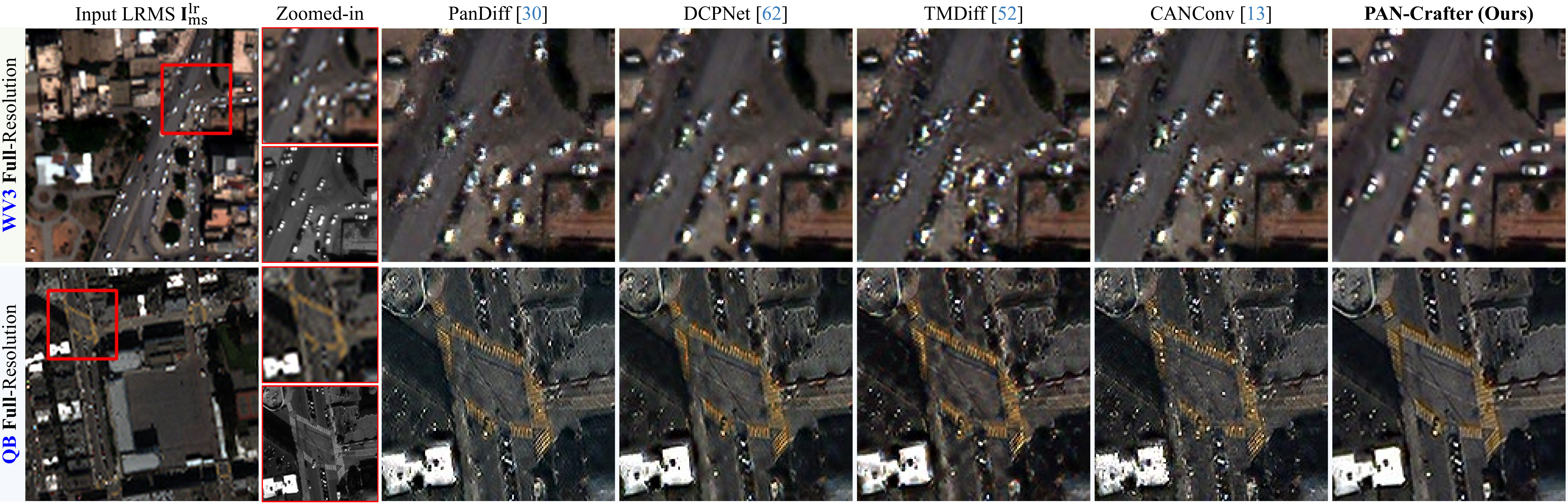}
  \vspace{-0.6cm}
  \caption{Visual comparison of PAN-Sharpening (PS) results on the WV3 and QB datasets at full-resolution. The leftmost column shows the input LRMS images, with {\color{red}{red boxes}} indicating zoomed-in regions for both LRMS and PAN images. Only our proposed PAN-Crafter is capable of generating high-quality images with clear edges around cars, buildings, and crosswalk lines, whereas previous methods tend to produce blurry and distorted results from misaligned input PAN and MS pair images.}
  \label{fig:full_result}
  \vspace{-0.2cm}
\end{figure*}

\begin{table*}[tbp]
\scriptsize
\centering
\resizebox{1.0\textwidth}{!}{ 
\def\arraystretch{1.2}
\begin{tabular}{l|l|ccc|cccccc|cc}
\Xhline{2\arrayrulewidth}
\multicolumn{2}{c|}{\textbf{WV3} Dataset} & \multicolumn{3}{c|}{Full-Resolution} & \multicolumn{6}{c|}{Reduced-Resolution} & \multirow{2}{*}{\makecell{Inference\\Time$\downarrow$ (s)}} & \multirow{2}{*}{\makecell{Memory$\downarrow$\\(GB)}} \\
\cline{1-11}
Methods & Publications & HQNR$\uparrow$ & $D_{s}$$\downarrow$ & $D_{\lambda}$$\downarrow$ & ERGAS$\downarrow$ & SCC$\uparrow$ & SAM$\downarrow$ & Q8$\uparrow$ & PSNR$\uparrow$ & SSIM$\uparrow$ \\
\hline
PanNet \cite{yang2017pannet} & ICCV 2017 & 0.918 & 0.049 & 0.035 & 2.538 & 0.979 & 3.402 & 0.913 & 36.148 & 0.966 & - & - \\
MSDCNN \cite{yuan2018multiscale} & JSTARS 2018 & 0.924 & 0.050 & 0.028 & 2.489 & 0.979 & 3.300 & 0.914 & 36.329 & 0.967 & - & - \\
FusionNet \cite{wu2021dynamic} & ICCV 2021 & 0.920 & 0.053 & 0.029 & 2.428 & 0.981 & 3.188 & 0.916 & 36.569 & 0.968 & - & - \\
LAGConv \cite{jin2022lagconv} & AAAI 2022 & 0.915 & 0.055 & 0.033 & 2.380 & 0.981 & 3.153 & 0.916 & 36.732 & 0.970 & {\color{red}{\textbf{0.004}}} & 3.281 \\
S2DBPN \cite{zhang2023spatial} & TGRS 2023 & 0.946 & 0.030 & 0.025 & 2.245 & 0.985 & 3.019 & 0.917 & 37.216 & 0.972 & 0.005 & 2.387 \\
PanDiff \cite{meng2023pandiff} & TGRS 2023 & 0.952 & 0.034 & {\color{red}{\textbf{0.014}}} & 2.276 & 0.984 & 3.058 & 0.913 & 37.029 & 0.971 & 2.955 & 2.328 \\
DCPNet \cite{zhang2024dcpnet} & TGRS 2024 & 0.923 & 0.036 & 0.043 & 2.301 & 0.984 & 3.083 & 0.915 & 37.009 & 0.972 & 0.109 & 7.213 \\
TMDiff \cite{xing2024empower} & TGRS 2024 & 0.924 & 0.059 & 0.018 & 2.151 & 0.986 & 2.885 & 0.915 & 37.477 & 0.973 & 9.997 & 9.910 \\
CANConv \cite{duan2024content} & CVPR 2024 & 0.951 & 0.030 & 0.020 & 2.163 & 0.985 & 2.927 & 0.918 & 37.441 & 0.973 & 0.451 & 2.713 \\
\textbf{PAN-Crafter} & - & {\color{red}{\textbf{0.958}}} & {\color{red}{\textbf{0.027}}} & 0.016 & {\color{red}{\textbf{2.040}}} & {\color{red}{\textbf{0.988}}} & {\color{red}{\textbf{2.787}}} & {\color{red}{\textbf{0.922}}} & {\color{red}{\textbf{37.956}}} & {\color{red}{\textbf{0.976}}} & 0.009 & {\color{red}{\textbf{1.711}}} \\
\Xhline{2\arrayrulewidth}
\end{tabular}}
\vspace{-0.2cm}
\caption{Quantitative comparison of deep learning-based PS methods on the WV3 dataset. \textbf{\color{red}{Red}} indicate the best performance in each metric. The inference time and memory usage are measured on a $256 \times 256 \times 8$ HRMS target at reduced-resolution.}
\vspace{-0.3cm}
\label{tab:wv3}
\end{table*}

\noindent To further enhance the MS feature representation with PAN’s structural information, alignment-aware attention allows $\mathbf{Q}$ to attend to $\mathbf{K}$-$\mathbf{V}$ pairs derived from a down-sampled PAN image $\mathbf{I}_{\text{pan}}^{\text{rep},\downarrow}$:
\begin{equation}
    \begin{split}
    & \left[\mathbf{K}_{\text{pan}} \mid \mathbf{V}_{\text{pan}}\right] = \mathsf{Conv}\left(\left[\mathbf{I}_{\text{pan}}^{\text{rep},\downarrow} \mid \mathbf{x}\right]\right),\\
    & \mathbf{x}_{\text{pan}} = \mathsf{LocalAttn}(\mathbf{Q}, \mathbf{K}_{\text{pan}}, \mathbf{V}_{\text{pan}}),\\
    \end{split}
\end{equation}

\noindent This process enables $\mathcal{P}_\theta$ to extract high-frequency details from PAN images while mitigating cross-modality misalignment. In $\mathsf{PAN}$ mode, $\mathcal{P}_\theta$ back-reconstructs $\mathbf{I}_{\text{pan}}^{\text{rep}}$, prioritizing the preservation of sharp spatial details while leveraging spatial information from MS images. Here, $\mathbf{Q}$ is constructed differently to reflect the modality shift. Instead of using $\mathbf{I}_{\text{ms}}^{\text{lr},\downarrow}$, the query is formed by concatenating the input feature $\mathbf{x}$ with $\mathbf{I}_{\text{pan}}^{\text{rep},\downarrow}$ as:

\begin{equation}
    \mathbf{Q} = \mathsf{Conv}\left(\left[\mathbf{I}_{\text{pan}}^{\text{rep},\downarrow} \mid \mathbf{x}\right]\right).
\end{equation}

\noindent The subsequent self-attention and alignment-aware attention operations mirror those in $\mathsf{MS}$ mode but in reverse, ensuring structural consistency in PAN back-reconstruction. Since PAN back-reconstruction serves as an auxiliary task, it reinforces spatial sharpness in HRMS prediction. By jointly leveraging self-attention for modality-consistent refinement and alignment-aware attention for cross-modality adaptation, CM3A effectively mitigates misalignment while preserving spectral fidelity and structural coherence, ultimately enhancing HRMS quality.
\section{Experiments}
\label{sec:experiment}

\subsection{Datasets}

We evaluate PAN-Crafter on four widely used PAN-sharpening datasets from PanCollection \cite{deng2022machine}: WorldView-3 (WV3), QuickBird (QB), GaoFen-2 (GF2), and WorldView-2 (WV2). For training, we use $64\times64\times1$ patches for PAN and $16\times16\times C_{\text{ms}}$ patches for MS, where $C_{\text{ms}}=4$ for QB and GF2, and $C_{\text{ms}}=8$ for WV3. Each satellite dataset has its own test set, which consists of reduced-resolution and full-resolution images. To further assess generalization, we evaluate PAN-Crafter on WV2, an unseen satellite dataset used exclusively for testing. WV2 serves as a zero-shot benchmark, measuring the model’s robustness to sensor variations without fine-tuning. The reduced-resolution test images have a PAN spatial size of $256\times256$, while the full-resolution test images have a higher PAN spatial size of $512\times512$. 

\begin{figure*}[tbp]
  \centering
  \includegraphics[width=1.0\textwidth]{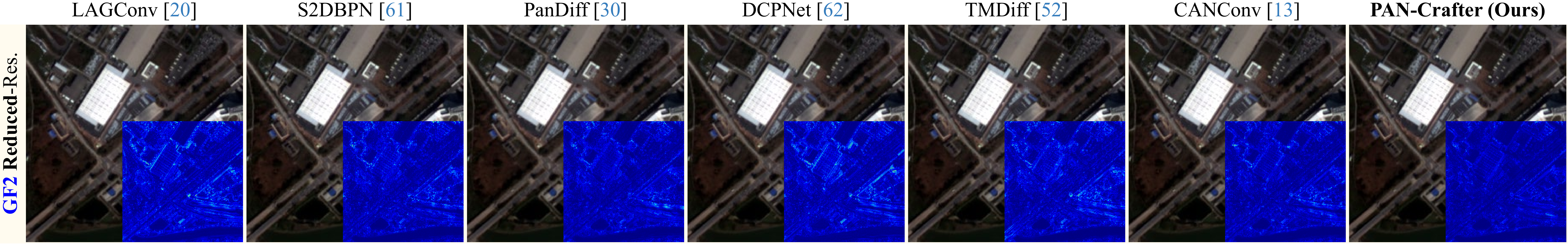}
  \vspace{-0.6cm}
  \caption{Visual comparison of PS results on the GF2 and QB datasets at reduced-resolution. The {\color{blue}{blue-colored}} insets represent error maps computed against the ground truth (GT), where brighter regions indicate higher reconstruction errors.}
  \label{fig:reduced_result}
  \vspace{-0.2cm}
\end{figure*}

\begin{table*}[tbp]
\scriptsize
\centering
\resizebox{1.0\textwidth}{!}{ 
\def\arraystretch{1.2}
\begin{tabular}{l|cc|cccc|cc|cccc}
\Xhline{2\arrayrulewidth}
\multirow{3}{*}{Methods} & \multicolumn{6}{c|}{\textbf{GF2} Dataset} & \multicolumn{6}{c}{\textbf{QB} Dataset}\\
\cline{2-13}
& \multicolumn{2}{c|}{Full-Resolution} & \multicolumn{4}{c|}{Reduced-Resolution} & \multicolumn{2}{c|}{Full-Resolution} & \multicolumn{4}{c}{Reduced-Resolution}\\
\cline{2-13}
& HQNR$\uparrow$ & $D_{s}$$\downarrow$ & ERGAS$\downarrow$ & SCC$\uparrow$ & SAM$\downarrow$ & PSNR$\uparrow$ & HQNR$\uparrow$ & $D_{s}$$\downarrow$ & ERGAS$\downarrow$ & SCC$\uparrow$ & SAM$\downarrow$ & PSNR$\uparrow$ \\
\hline
PanNet \cite{yang2017pannet} & 0.929 & 0.052 & 1.038 & 0.975 & 1.050 & 39.197 & 0.851 & 0.092 & 4.856 & 0.966 & 5.273 & 35.563 \\
MSDCNN \cite{yuan2018multiscale} & 0.898 & 0.079 & 0.862 & 0.983 & 0.946 & 40.730 & 0.888 & 0.058 & 4.074 & 0.977 & 4.828 & 37.040 \\
FusionNet \cite{wu2021dynamic} & 0.865 & 0.105 & 0.960 & 0.980 & 0.971 & 39.866 & 0.853 & 0.079 & 4.183 & 0.975 & 4.892 & 36.821 \\
LAGConv \cite{jin2022lagconv} & 0.895 & 0.078 & 0.816 & 0.985 & 0.886 & 41.147 & 0.892 & {\color{red}{\textbf{0.035}}} & 3.845 & 0.980 & 4.682 & 37.565 \\
S2DBPN \cite{zhang2023spatial} & 0.935 & 0.046 & 0.686 & 0.990 & 0.772 & 42.686 & 0.908 & 0.036 & 3.956 & 0.980 & 4.849 & 37.314 \\
PanDiff \cite{meng2023pandiff} & 0.936 & 0.045 & 0.674 & 0.990 & 0.767 & 42.827 & 0.919 & 0.055 & 3.723 & 0.982 & 4.611 & 37.842 \\
DCPNet \cite{zhang2024dcpnet} & 0.953 & 0.024 & 0.724 & 0.988 & 0.806 & 42.312 & 0.880 & 0.073 & 3.618 & 0.983 & {\color{red}{\textbf{4.420}}} & 38.079 \\
TMDiff \cite{xing2024empower} & 0.942 & 0.030 & 0.754 & 0.988 & 0.764 & 41.896 & 0.901 & 0.068 & 3.804 & 0.981 & 4.627 & 37.642 \\
CANConv \cite{duan2024content} & 0.919 & 0.063 & 0.653 & 0.991 & 0.722 & 43.166 & 0.893 & 0.070 & 3.740 & 0.982 & 4.554 & 37.795 \\
\hline
\textbf{PAN-Crafter} & {\color{red}{\textbf{0.964}}} & {\color{red}{\textbf{0.017}}} & {\color{red}{\textbf{0.522}}} & {\color{red}{\textbf{0.994}}} & {\color{red}{\textbf{0.596}}} & {\color{red}{\textbf{45.076}}} & {\color{red}{\textbf{0.920}}} & 0.039 & {\color{red}{\textbf{3.570}}} & {\color{red}{\textbf{0.984}}} & 4.426 & {\color{red}{\textbf{38.195}}} \\
\Xhline{2\arrayrulewidth}
\end{tabular}}
\vspace{-0.2cm}
\caption{Quantitative comparison of deep learning-based PS methods on the GF2 and QB datasets. \textbf{\color{red}{Red}} indicates the best performance.}
\label{tab:gf2_qb}
\vspace{-0.4cm}
\end{table*}

\subsection{Experiment Details}

We implement PAN-Crafter in PyTorch \cite{Pytorch} and conduct all experiments on a single NVIDIA GeForce RTX 3090 GPU. Each model is trained for 50,000 iterations with a 100-step warmup period. We use AdamW optimizer \cite{AdamW} with an initial learning rate of $1 \times 10^{-4}$, a weight decay of 0.01, and a cosine-annealing scheduler \cite{CosineAnneal} to progressively reduce the learning rate. The batch size is set to 48, but for MARs loss computation, the batch is duplicated across $\mathsf{MS}$ mode and $\mathsf{PAN}$ mode, resulting in an effective batch size of 96. We empirically set the loss weight to $\lambda = 1.0$ and the local attention kernel size to $k = 3$. All feature dimensions are fixed to $C = 128$. Standard data augmentation techniques, including random horizontal/vertical flips, 90-degree rotations, and random cropping, are applied to improve generalization. To ensure reproducibility, we fix the random seed to 2,025 across all experiments. We evaluate PAN-Crafter in terms of ERGAS \cite{wald2000quality}, SCC \cite{garzelli2009hypercomplex}, SAM \cite{yuhas1992discrimination}, Q4/Q8 \cite{vivone2014critical}, PSNR \cite{wang2004image}, and SSIM \cite{wang2004image} metrics for reduced-resolution datasets, while HQNR \cite{vivone2020new}, $D_S$, and $D_{\lambda}$ metrics are used for full-resolution datasets.

\begin{figure*}[tbp]
  \centering
  \includegraphics[width=1.0\textwidth]{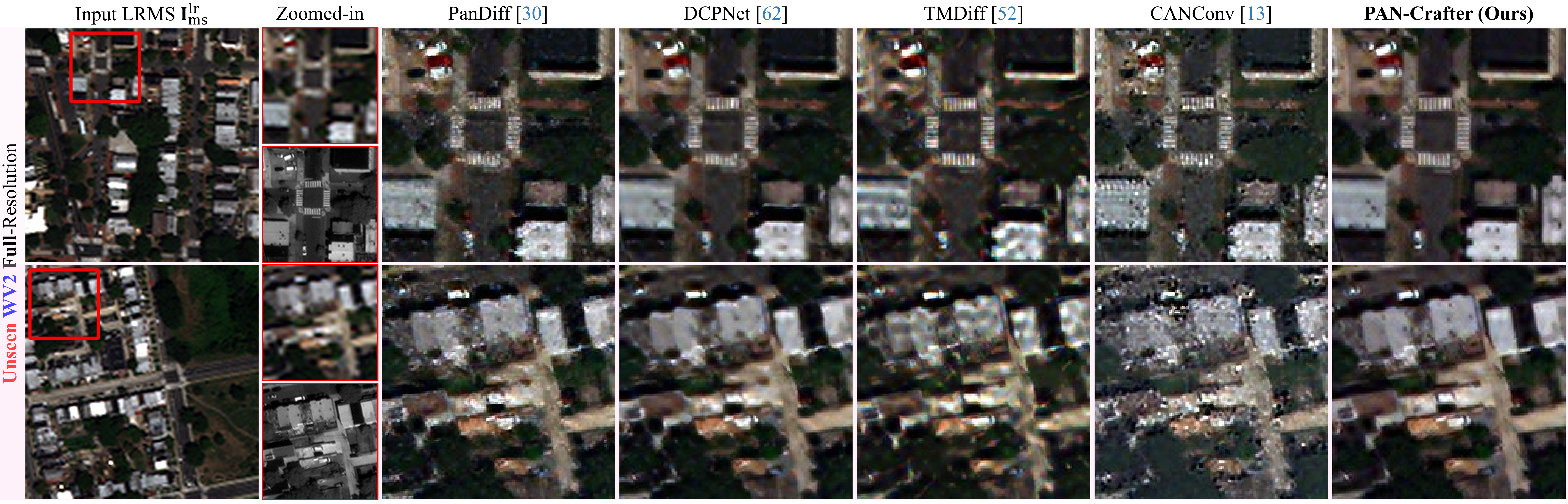}
  \vspace{-0.6cm}
  \caption{Visual comparison of PS results on the unseen WV2 dataset at full-resolution. The leftmost column shows the input LRMS image, with {\color{red}red boxes} indicating zoomed-in regions for both LRMS and PAN images. Since WV2 is not included in the training phase, this evaluation represents a real-world zero-shot setting, assessing the generalization capability of PS models. Our PAN-Crafter produces sharper details with fewer distortions compared to existing methods, demonstrating superior cross-satellite robustness.}
  \label{fig:unseen_result}
  \vspace{-0.4cm}
\end{figure*}

\subsection{Experimental Results}

We computed all evaluation metrics using the official PanCollection \cite{deng2022machine} repository to ensure standardized measurement. To ensure a fair and comprehensive evaluation, we utilized the official implementations of the compared methods whenever available. 

\noindent\textbf{Qualitative comparison.} We qualitatively compare our PAN-Crafter against very recent PS methods on both full- (Fig.~\ref{fig:first}, Fig.~\ref{fig:full_result}) and reduced-resolution (Fig.~\ref{fig:reduced_result}) datasets. Previous methods often produce blurry artifacts, double edges, or spectral distortions due to their inability to handle cross-modality misalignment. In contrast, PAN-Crafter effectively preserves fine details, ensuring sharper edges and clearer textures by leveraging CM3A attention for explicit cross-modality alignment. Beyond direct feature alignment, MARs further enhances robustness by leveraging $\mathsf{PAN}$ mode as an auxiliary self-supervision mechanism, encouraging the network to distill sharper structural details into the final HRMS output. The synergy of MARs and CM3A allows PAN-Crafter to generate high-quality HRMS images with superior spatial and spectral integrity, demonstrating its effectiveness in mitigating misalignment.

\noindent\textbf{Quantitative evaluation.} We compare PAN-Crafter against deep learning-based PS methods on WV3, GF2, and QB datasets. As shown in Table~\ref{tab:wv3} and  Table~\ref{tab:gf2_qb}, our PAN-Crafter consistently outperforms existing approaches across most evaluation metrics while maintaining low memory consumption and fast inference time. Notably, our method surpasses diffusion-based models (PanDiff \cite{meng2023pandiff} and TMDiff \cite{xing2024empower}) in both full- and reduced-resolution datasets, demonstrating that our CM3A module effectively handles local misalignment without the computational burden of iterative diffusion process. Specifically, PAN-Crafter achieves 328.33$\times$ and 1110.78$\times$ faster inference time compared to PanDiff and TMDiff, respectively. Additionally, compared to CANConv \cite{duan2024content}, which relies on k-means clustering \cite{ding2015yinyang} for kernel generation, PAN-Crafter is 50.11$\times$ faster, highlighting the efficiency of our locally adaptive alignment strategy. Notably, on the WV3 dataset, the $D_{\lambda}$ score is slightly lower than other methods. This is because our method aligns the generated HRMS image to the LRMS image rather than the PAN image using CM3A in $\mathsf{MS}$ mode, ensuring spectral fidelity at the cost of a lower PAN-HRMS correlation. For the QB dataset, which is known to be the most challenging due to its higher spectral distortion and complex scene variations, PAN-Crafter exhibits slightly lower $D_s$ and SAM scores compared to some methods. However, it still achieves the best overall performance across both full- and reduced-resolution evaluations, confirming its robustness to diverse satellite imagery.

\begin{table}[tbp]
\scriptsize
\centering
\resizebox{1.0\columnwidth}{!}{ 
\def\arraystretch{1.2}
\begin{tabular}{l|c|cccc}
\Xhline{2\arrayrulewidth}
\multirow{2}{*}{Methods} & \multicolumn{5}{c}{\textbf{WV2} Dataset (\textbf{Unseen} satellite dataset)} \\
\cline{2-6}
& HQNR$\uparrow$ & ERGAS$\downarrow$ & SCC$\uparrow$ & SAM$\downarrow$ & PSNR$\uparrow$ \\
\hline
PanNet \cite{yang2017pannet} & 0.875 & 5.481 & 0.876 & 7.040 & 27.120 \\
MSDCNN \cite{yuan2018multiscale} & 0.862 & 4.930 & 0.905 & 5.898 & 27.901 \\
FusionNet \cite{wu2021dynamic} & 0.862 & 5.100 & 0.902 & 6.118 & 27.616 \\
LAGConv \cite{jin2022lagconv} & 0.902 & 5.133 & 0.885 & 6.094 & 27.525 \\
S2DBPN \cite{zhang2023spatial} & 0.813 & 5.703 & 0.915 & 7.063 & 26.748 \\
PanDiff \cite{meng2023pandiff} & 0.932 & 4.291 & 0.916 & 5.430 & 28.964 \\
DCPNet \cite{zhang2024dcpnet} & 0.797 & 5.507 & {\color{red}{\textbf{0.931}}} & 10.174 & 27.063 \\
TMDiff \cite{xing2024empower} & 0.874 & 5.157 & 0.875 & 6.087 & 27.473 \\
CANConv \cite{duan2024content} & 0.876 & 4.328 & 0.918 & 5.481 & 29.005 \\
\textbf{PAN-Crafter} & {\color{red}{\textbf{0.942}}} & {\color{red}{\textbf{4.169}}} & 0.924 &  {\color{red}{\textbf{5.078}}} & {\color{red}{\textbf{29.276}}} \\
\Xhline{2\arrayrulewidth}
\end{tabular}}
\vspace{-0.2cm}
\caption{Quantitative comparison of deep learning-based PS methods on the unseen WV2 satellite dataset. All models are trained on WV3 and evaluated on WV2 to assess real-world generalization.}
\label{tab:wv2}
\vspace{-0.4cm}
\end{table}

\noindent\textbf{Generalization on unseen satellite dataset.} To assess the generalization capability of PAN-Crafter, we conduct fully zero-shot evaluations on the WV2 dataset, an unseen satellite dataset not included in the training phase. As shown in Table~\ref{tab:wv2} and Fig.~\ref{fig:unseen_result}, all models are trained on WV3 and directly tested on WV2 without any fine-tuning. While existing methods suffer from performance degradation due to domain shifts and sensor variations, PAN-Crafter demonstrates superior generalization ability, achieving the best performance across all key metrics. The strong generalization stems from our modality-consistent alignment strategy, which explicitly mitigates cross-modality misalignment without relying on dataset-specific priors. These results highlight the robustness of our framework in real-world PS scenarios, where test-time adaptation is often infeasible.

\subsection{Ablation Studies}

\noindent\textbf{Impact of MARs mode.} As shown in Table~\ref{tab:ablation}, incorporating MARs improves all evalution metric, particularly enhancing SAM and PSNR scores. This highlights the effectiveness of PAN as an auxiliary self-supervision signal, allowing the network to learn sharper spatial details while maintaining spectral fidelity. Despite a slight increase in memory usage, MARs provides a substantial performance gain, justifying its inclusion in our framework.

\noindent\textbf{Analysis of CM3A.} In Table~\ref{tab:ablation}, we further examine the impact of our Cross-Modality Alignment-Aware Attention (CM3A). Without CM3A, the model struggles to mitigate cross-modality misalignment, leading to degraded spectral and spatial consistency. The results show that CM3A significantly improves HQNR and ERGAS scores while reducing SAM, demonstrating its effectiveness in aligning MS textures with the corresponding PAN structures.

\begin{table}[tbp]
    \scriptsize
    \centering
    \resizebox{1.0\columnwidth}{!}{
    \def\arraystretch{1.2}
    \begin{tabular} {c|c|c|ccc|cc}
        \Xhline{2\arrayrulewidth}
        \multirow{2}{*}{CM3A} &  \multirow{2}{*}{MARs} & \multicolumn{6}{c}{\textbf{WV3} Dataset} \\
        \cline{3-8}
        & & HQNR$\uparrow$ & ERGAS$\downarrow$ & SAM$\downarrow$ & PSNR$\uparrow$ & Time$\downarrow$ & Memory$\downarrow$ \\
        \hline
        & & 0.948 & 2.232 & 2.980 & 37.245 & 0.006 & 1.537 \\
        \checkmark & & 0.949 & 2.212 & 2.970 & 37.285 & 0.007 & 1.556 \\
        & \checkmark & 0.956 & 2.122 & 2.873 & 37.602 & 0.009 & 1.701 \\
        \checkmark & \checkmark & {\color{red}{\textbf{0.958}}} & {\color{red}{\textbf{2.040}}} & {\color{red}{\textbf{2.787}}} & {\color{red}{\textbf{37.956}}} & 0.009 & 1.711 \\
        \Xhline{2\arrayrulewidth}
    \end{tabular}}
    \vspace{-0.2cm}
    \caption{Ablation studies on CM3A and MARs on the WV3 dataset. Notably, the combination of both components achieves the best performance, highlighting their synergistic effect in jointly refining spatial and spectral consistency.}
  \label{tab:ablation}
  \vspace{-0.4cm}
\end{table}

\section{Conclusion}
\label{sec:conclusion}

We introduce PAN-Crafter, a modality-consistent alignment framework for PAN-sharpening that explicitly addresses cross-modality misalignment. MARs enables joint HRMS and PAN reconstruction, leveraging PAN’s high-frequency details as auxiliary self-supervision, while CM3A ensures bidirectional alignment between MS and the corresponding PAN images. PAN-Crafter achieves SOTA performance across multiple benchmarks, preserving fine details and spectral integrity. It also generalizes well to unseen satellite data, demonstrating strong zero-shot robustness. With superior efficiency in inference speed and memory usage, PAN-Crafter offers a practical and scalable solution for real-world remote sensing applications.

\vspace{0.2cm}
\noindent \textbf{Acknowledgement.}\quad  This work was supported by National Research Foundation of Korea (NRF) grant funded by the Korean Government [Ministry of Science and ICT (Information and Communications Technology)] (Project Number: RS2024-00338513, Project Title: AI-based Computer Vision Study for Satellite Image Processing and Analysis, 100\%).

\clearpage
\maketitlesupplementary
\appendix

\section{Additional Discussions on Results}

\subsection{Additional Qualitative Comparisons}

\begin{figure*}[tbp]
  \centering
  \includegraphics[width=1.0\textwidth]{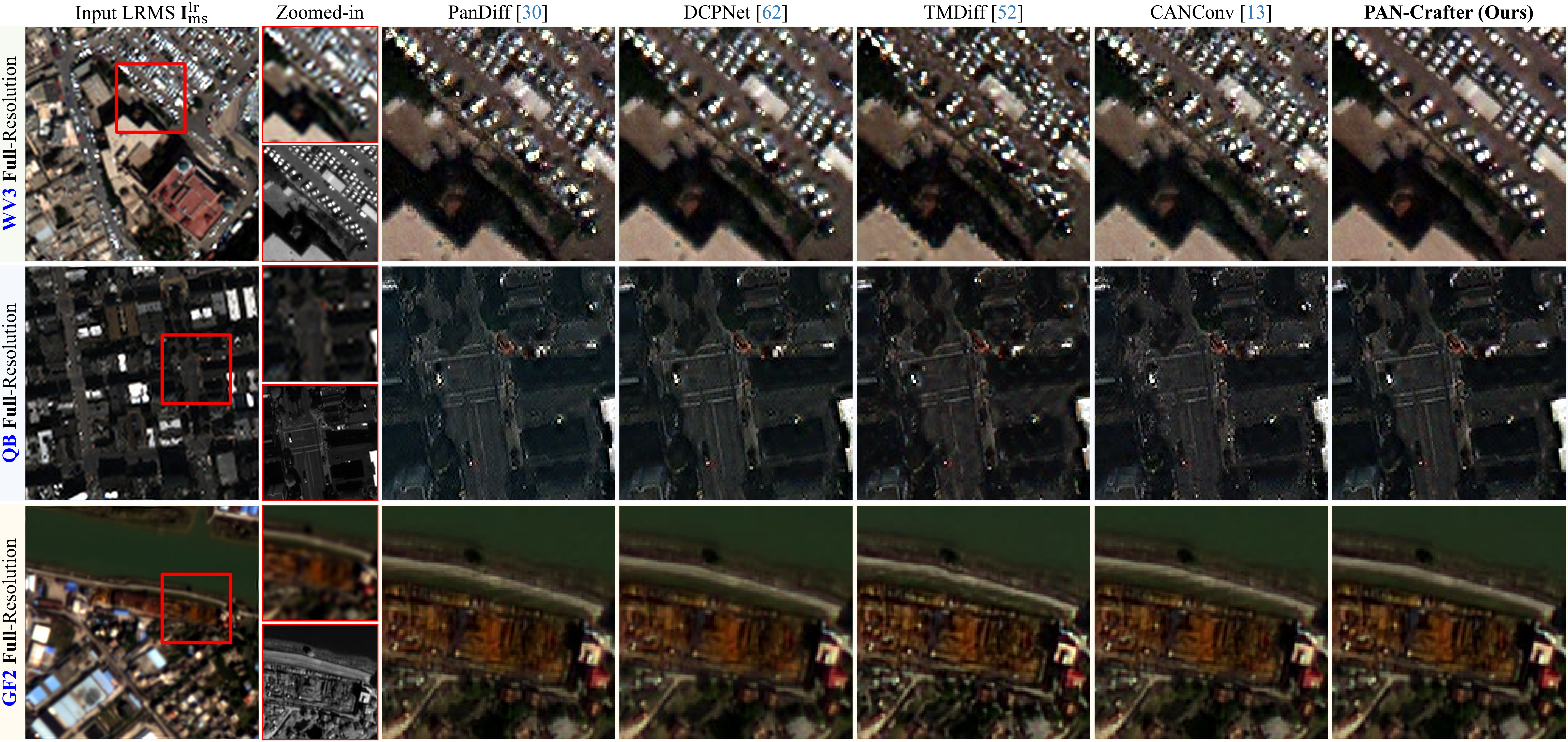}
  \caption{Visual comparison of PAN-Sharpening (PS) results on the WV3, QB, and GF2 datasets at full-resolution. The leftmost column shows the input LRMS images, with {\color{red}{red boxes}} indicating zoomed-in regions for both LRMS and PAN images. Our PAN-Crafter method generates pan-sharpened images with minimal artifacts, particularly around buildings and vehicles, whereas other methods frequently produce blurry or distorted outputs.}
  \label{fig:supple_full_1}
\end{figure*}

\begin{figure*}[tbp]
  \centering
  \includegraphics[width=1.0\textwidth]{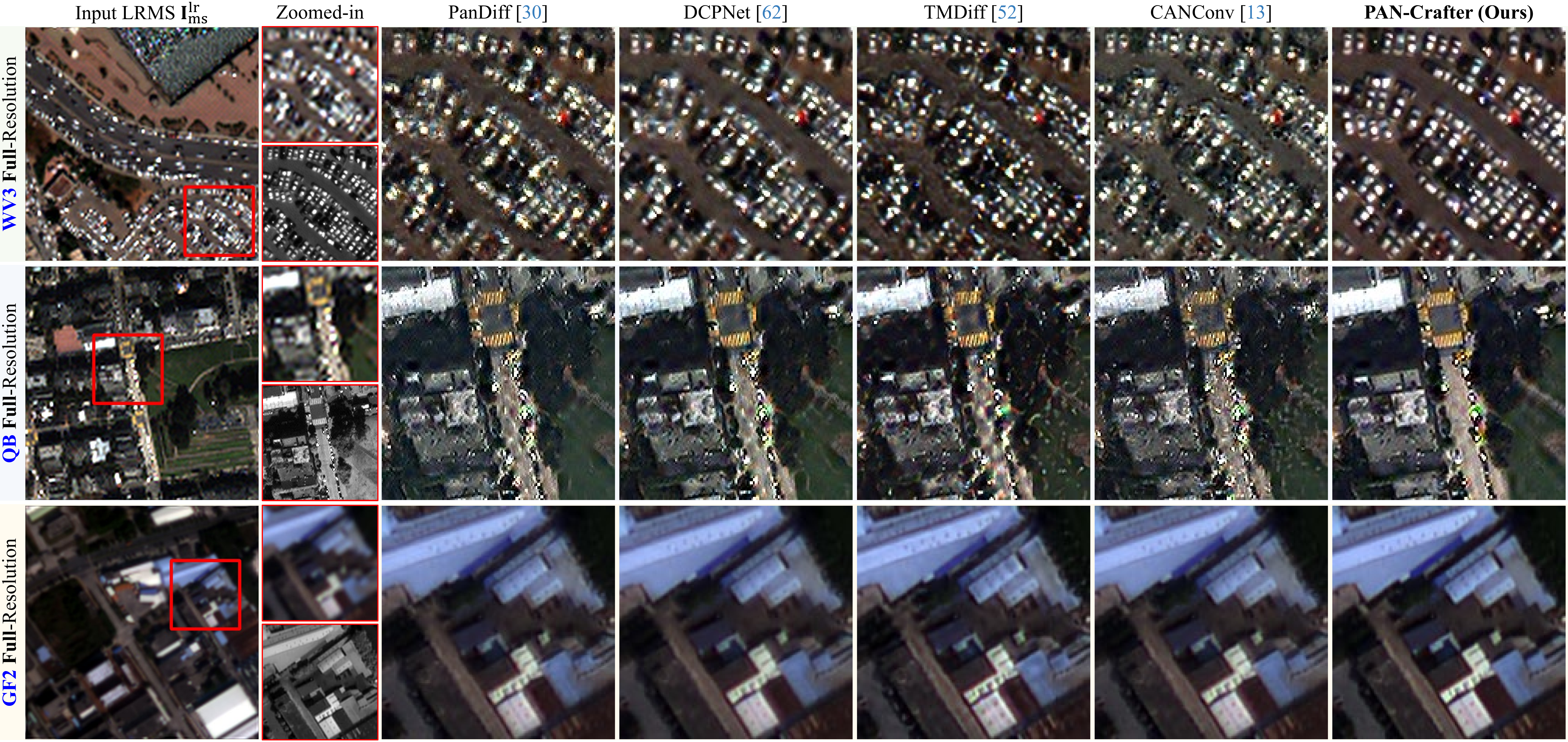}
  \caption{Visual comparison of PAN-Sharpening (PS) results on the WV3, QB, and GF2 datasets at full-resolution. The leftmost column shows the input LRMS images, with {\color{red}{red boxes}} indicating zoomed-in regions for both LRMS and PAN images. Our PAN-Crafter method generates high-quality of pan-sharpened images with minimal artifacts, particularly around vehicles and crosswalks, whereas other methods frequently produce blurry or distorted outputs.}
  \label{fig:supple_full_2}
\end{figure*}

\begin{figure*}[tbp]
  \centering
  \includegraphics[width=1.0\textwidth]{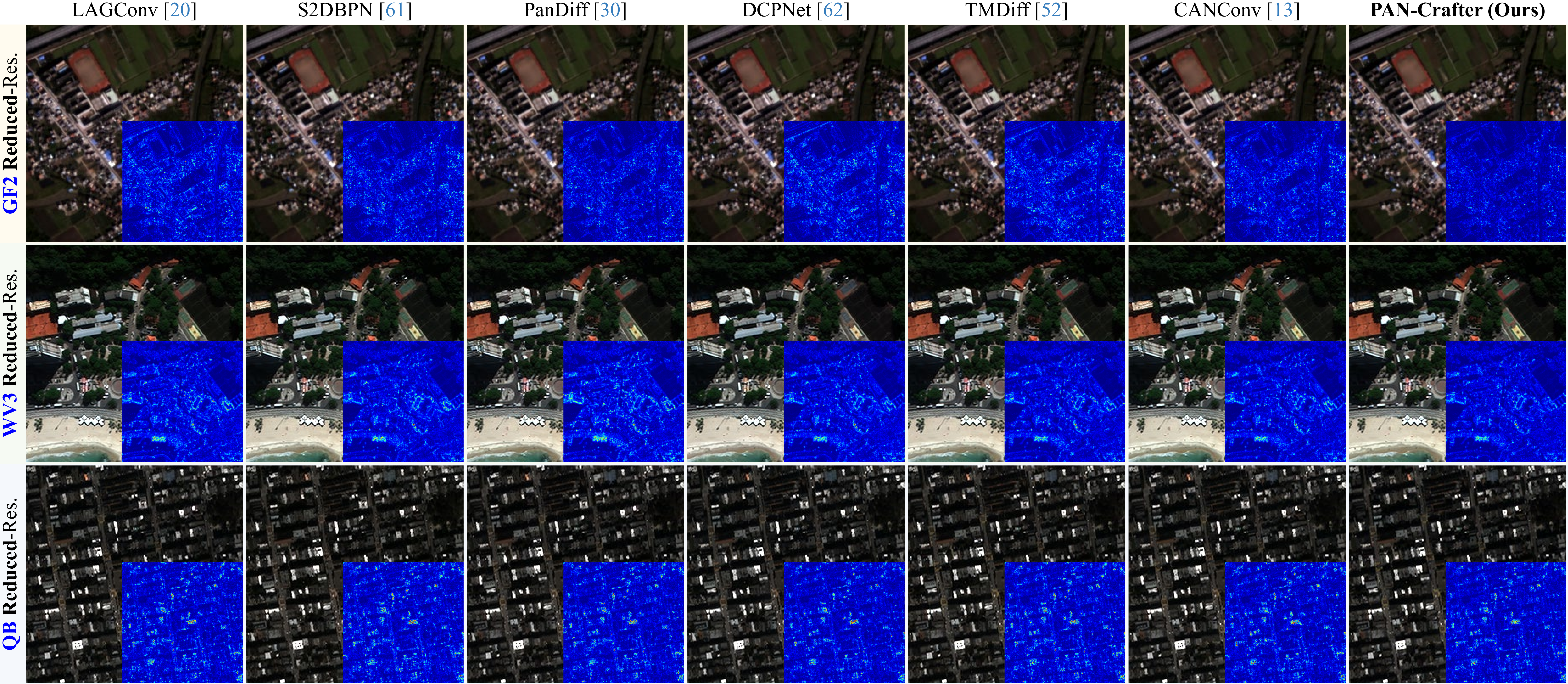}
  \caption{Visual comparison of PS results on the GF2 and QB datasets at reduced-resolution. The {\color{blue}{blue-colored}} insets represent error maps computed against the ground truth (GT), where brighter regions indicate higher reconstruction errors.}
  \label{fig:supple_reduced_1}
\end{figure*}

\begin{figure*}[tbp]
  \centering
  \includegraphics[width=1.0\textwidth]{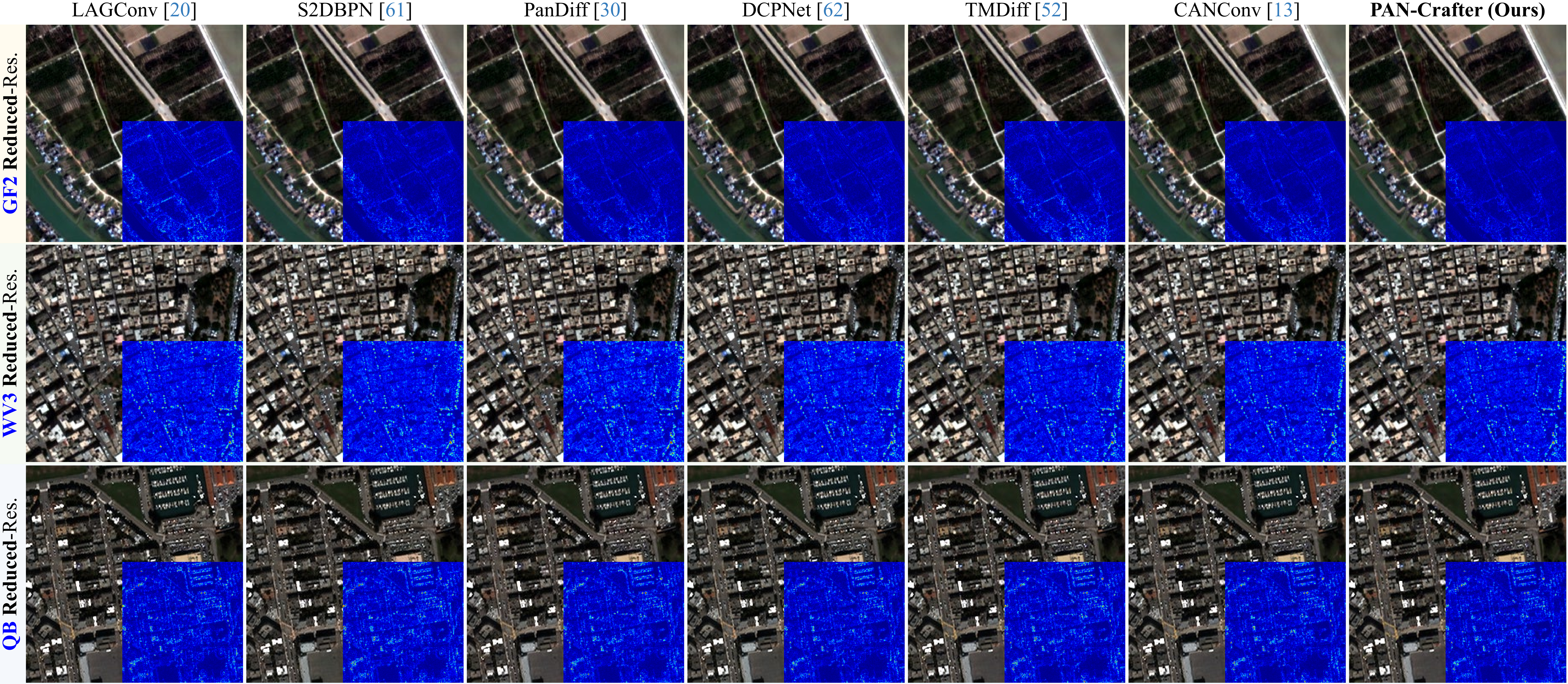}
  \caption{Visual comparison of PS results on the GF2 and QB datasets at reduced-resolution. The {\color{blue}{blue-colored}} insets represent error maps computed against the ground truth (GT), where brighter regions indicate higher reconstruction errors.}
  \label{fig:supple_reduced_2}
\end{figure*}

Fig.~\ref{fig:supple_full_1} and Fig.~\ref{fig:supple_full_2} provide additional qualitative comparisons of PS results on the WorldView-3 (WV3), QuickBird (QB), and GaoFen-2 (GF2) datasets \cite{deng2022machine} at full-resolution. Fig.~\ref{fig:supple_reduced_1} and Fig.~\ref{fig:supple_reduced_2} provide additional qualitative comparisons of PS results on the WV3, QB, GF2 datasets at reduced-resolution. Our PAN-Crafter consistently generates pan-sharpened images with minimal artifacts, preserving fine details around buildings and vehicles, whereas existing methods often produce blurring or structural distortions.

\subsection{Additional Quantitative Evaluation}

To provide a more comprehensive analysis, we present extended quantitative evaluations in the \textit{Supplementary Material}. Table~\ref{tab:wv3_detail}, Table~\ref{tab:gf2_detail}, and Table~\ref{tab:qb_detail} provide detailed results on the WV3, GF2, and QB datasets, respectively. The (i) metrics, (ii) full-/no-reference, and (iii) Wald’s evaluation protocol \cite{vivone2014critical, thomas2008synthesis} used for evaluations are quite commonly known for PS restoration in remote sensing. We confirm that all comparison methods were trained using their official codebases and were evaluated, following the same protocol \cite{vivone2014critical, thomas2008synthesis} used in the prior work \cite{duan2024content, meng2023pandiff, xing2024empower, zhang2024dcpnet}. To ensure fairness, we applied the same data splits \cite{deng2022machine}, random seeds, data augmentations. PAN-Crafter consistently achieves strong performance across various evaluation metrics, further demonstrating its effectiveness in preserving both spatial and spectral fidelity. These extended results reinforce the robustness of our approach across different datasets and imaging conditions.

\subsection{Generalization on Unseen Satellite Dataset}

\begin{figure*}[tbp]
  \centering
  \includegraphics[width=1.0\textwidth]{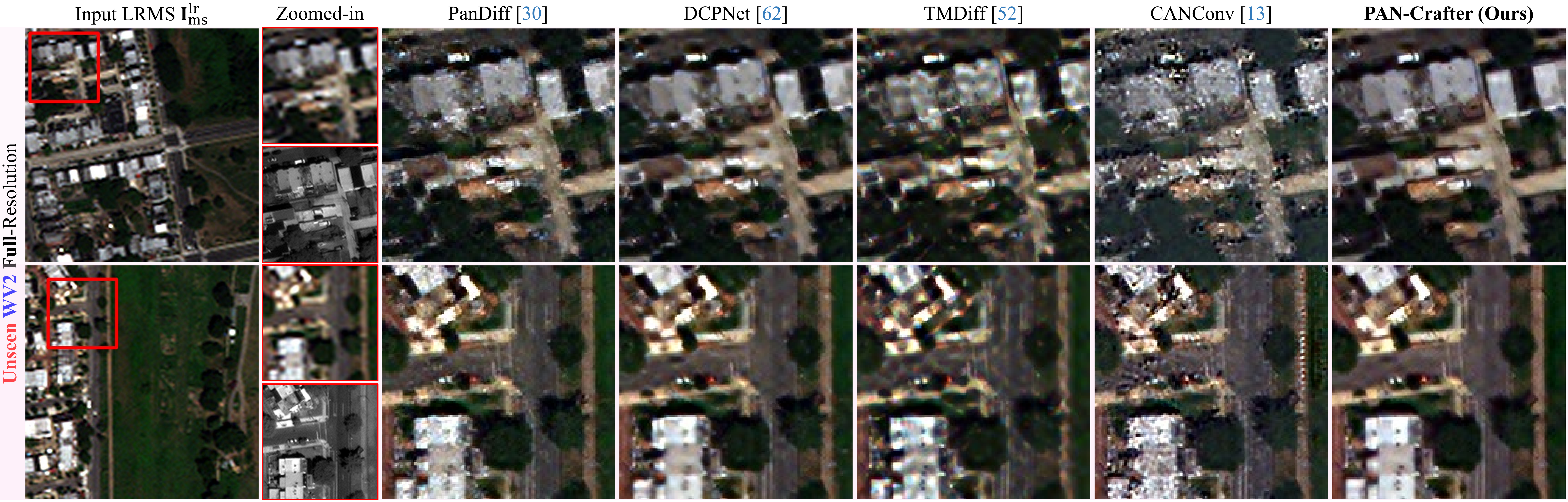}
  \caption{Visual comparison of PS results on the unseen WV2 dataset at full-resolution. The leftmost column shows the input LRMS image, with {\color{red}red boxes} indicating zoomed-in regions for both LRMS and PAN images. Since WV2 is not included in the training phase, this evaluation represents a real-world zero-shot setting, assessing the generalization capability of PS models. Our proposed PAN-Crafter significantly outperforms the existing methods by effectively preserving both fine structural and spectral details of the input MS and PAN images}
  \label{fig:supple_unseen}
\end{figure*}

To further evaluate the zero-shot generalization capability of PAN-Crafter, we provide additional quantitative and qualitative results on the unseen WorldView-2 (WV2) dataset \cite{deng2022machine}. Table~\ref{tab:wv2_detail} and Fig.~\ref{fig:supple_unseen} present quantitative and qualitative results, respectively. Despite not being trained on WV2, PAN-Crafter outperforms existing methods in both spatial and spectral fidelity, demonstrating its robustness to cross-sensor variations. The results highlight the effectiveness of our cross-modality alignment strategy, enabling strong generalization without requiring additional fine-tuning.

\subsection{Computational Complexity}

Efficiency is a critical factor in PS applications, particularly for real-time and large-scale remote sensing tasks. We evaluate the computational complexity of PAN-Crafter against state-of-the-art methods in terms of inference time, memory consumption, FLOPs, and the number of parameters, as summarized in Table~\ref{tab:complex}. Our PAN-Crafter achieves a significant speedup over diffusion-based models, with over 1110.78$\times$ faster inference time compared to TMDiff \cite{xing2024empower} and over 328.33$\times$ faster than PanDiff \cite{meng2023pandiff}, demonstrating the efficiency of our attention-based alignment mechanism. Compared to CANConv \cite{duan2024content}, which utilizes k-means clustering \cite{ding2015yinyang} for spatial adaptation, PAN-Crafter achieves 50.11$\times$ faster inference while maintaining competitive reconstruction quality.

\section{Limitations}

\subsection{Misalignment between multi-spectral bands}

Our method addresses cross-modality misalignment but does not explicitly handle misalignment between multi-spectral bands. A potential solution is to apply depth-wise separable convolutional layers in CM3A for MS feature projection, preventing information mixing across spectral bands.

\section{Further Ablation Studies}

\subsection{Ablation studies on MARs and CM3A} 

Table~\ref{tab:suppe_ablation_wv3}, Table~\ref{tab:suppe_ablation_gf2}, and Table~\ref{tab:suppe_ablation_qb} present extended ablation studies on MARs and CM3A across WV3, GF2, and QB datasets. The results demonstrate the significant impact of MARs, which consistently improves both spatial and spectral fidelity by leveraging auxiliary PAN self-supervision. While CM3A alone provides only marginal benefits, its effectiveness is significantly amplified when combined with MARs. The bidirectional interaction between PAN and MS reconstruction in MARs enables CM3A to refine cross-modality alignment more effectively, leading to a synergistic enhancement in both spatial consistency and spectral preservation. These findings further validate the importance of jointly leveraging MARs and CM3A for robust PAN-sharpening.

\noindent\textbf{Ablation settings.} We clarify the ablation setups for the main components: (i) without MARs -- we remove the \(\mathsf{PAN}\) mode entirely, including all learnable parameters related to modality switching (i.e., \(\bm{\alpha}\), \(\bm{\beta}\), and \(\bm{\gamma}\)). This turns the architecture into a single-task (PS) network; (ii) without CM3A -- we remove the concatenated original inputs (\(\mathbf{I}_{\text{ms}}^{\text{lr}, \downarrow}, \mathbf{I}_{\text{pan}}^{\text{rep}, \downarrow}\)) from the attention block, disabling cross-modality conditioning in the alignment mechanism.

\subsection{Additional ablation studies}

Additional component-wise ablations on the WV3, GF2, and QB datasets were done: (i) without modulation parameters (\(\bm{\beta}\), \(\bm{\gamma}\)) -- the modulation is not applied to the feature maps (Table~\ref{tab:suppe_ablation_1}); (ii) without combination parameter (\(\bm{\alpha}\)) -- we eliminate the learnable fusion weight between features (Table~\ref{tab:suppe_ablation_1}); (iii) varying local attention window size \(k\) in CM3A (Table~\ref{tab:suppe_ablation_2}); (iv) two-stage learning with pretrain on the PAN back-reconstruction and finetune for PS (Table~\ref{tab:suppe_ablation_3}).

\subsection{Justification of ablation results}

The reason that U-Net without MAR and CM3A is superior to existing methods is that our multi-scale window-based local attention \cite{pan2023slide} in U-Net is still effective to constitute a strong baseline. We additionally ablated this component and its result can be seen in the (Table~\ref{tab:suppe_ablation_4}). Without it (replacement of the local attention layer with convolution layer in the baseline model), the performance drop is significant.

\section{Discussion on Various Cross-Attention Approaches}

While the two prior works employ cross-attention in multi-frame restoration \cite{Zn5r_1} and reference-based SR \cite{Zn5r_2}, their setups are substantially different from ours. Siamtrans \cite{Zn5r_1} applies cross-attention after warping adjacent video frames to a query frame, assuming strong temporal correlation and accurate alignment. Similarly, TTSR \cite{Zn5r_2} uses global cross-attention between an LR image and a semantically unrelated HR reference. \cite{Zn5r_1, Zn5r_2} rely on global attention \cite{vaswani2017attention, dosovitskiy2020image}, which is computationally expensive and less suitable for local misalignment patterns. In contrast, our CM3A module is specially tailored for PS, where PAN and MS images are often not significantly misaligned and share similar spatial structures. To effectively handle this, we introduce a novel MARs-mode-dependent local cross-/self-attention. Also, we replace fixed positional embeddings (PE) with down-sampled original images concatenated to the attention inputs to implicitly learn the relative misalignment between modalities. The distinction between \cite{Zn5r_1, Zn5r_2} and our CM3A is summarized as Table~\ref{tab:novelty}.

\section{Local Attention Mechanisms}

Given a query feature $ \mathbf{Q} \in \mathbb{R}^{H \times W \times C} $ and key-value pairs $ \mathbf{K}, \mathbf{V} \in \mathbb{R}^{H \times W \times C} $, Local Attention fuction ($\mathsf{LocalAttn}$) \cite{pan2023slide} computes attention scores within the $ k \times k $ local receptive field as follows:

\begin{equation}
    \begin{split}
    &\mathsf{Attn}_{i,j,m,n} = \mathbf{Q}_{i,j}\mathbf{K}_{i+m,j+n},\\
    &\mathsf{Attn} \leftarrow \mathsf{SoftMax}\left(\mathsf{Attn} / \sqrt{C}\right),\\
    &\mathsf{LocalAttn}(\mathbf{Q}, \mathbf{K}, \mathbf{V})_{i,j}\\
    &= \sum_{m=-k^{\prime}}^{k^{\prime}} \sum_{n=-k^{\prime}}^{k^{\prime}}\mathsf{Attn}_{i,j,m,n}\mathbf{V}_{i+m,j+n},\\
    \end{split}
\end{equation}

\noindent where $\mathsf{Attn} \in \mathbb{R}^{H \times W \times k \times k}$ is the attention score, and $\mathsf{SoftMax}$ is applied along the last two dimensions.

\noindent \textbf{Computational complexity analysis.} The computational complexity of a global self-attention layer for a feature map $\mathbf{x}$ of size $ (H, W, C) $ is:

\begin{equation} \mathcal{O} (2(HW)^2 C), \end{equation}

\noindent due to pairwise interactions across all spatial locations. In contrast, CM3A leverages local attention with a fixed receptive field size of $ k \times k $, reducing the complexity to:

\begin{equation} \mathcal{O} (2 (HW) k^2 C). \end{equation}

\noindent Since $ k^2 \ll HW $, our approach significantly reduces computational overhead while maintaining effective cross-modality feature alignment. By restricting attention to local neighborhoods, CM3A balances efficiency with the ability to capture localized structural discrepancies between PAN and MS images.

\clearpage

\begin{table*}[tbp]
    \scriptsize
    \centering
    \resizebox{0.7\textwidth}{!}{
    \def\arraystretch{1.2}
    \begin{tabular}{l|l|l|l}
        \Xhline{2\arrayrulewidth}
        Methods & Tasks & Attn. types & Characteristics \\ 
        \hline
        Siamtrans \cite{Zn5r_1} & Multi-frame restoration & Global & Only cross-attention with PE \\
        \hline
        TTSR \cite{Zn5r_2} & Reference-based SR & Global & Only cross-attention with PE \\
        \hline
        \textbf{Ours} & PAN-sharpening & Local & MARs-mode-dependent cross-/self-attention \\
        \Xhline{2\arrayrulewidth}
    \end{tabular}}
    \caption{Comparison with ours CM3A with existing cross-attention methods.}
    \label{tab:novelty}
\end{table*}

\begin{table*}[tbp]
    \scriptsize
    \centering
    \resizebox{1.0\textwidth}{!}{ 
    \def\arraystretch{1.2}
    \begin{tabular}{l|ccc|cccccc}
    \Xhline{2\arrayrulewidth}
    \textbf{WV3} Dataset & \multicolumn{3}{c|}{Full-Resolution} & \multicolumn{6}{c}{Reduced-Resolution} \\
    \hline
    Methods & HQNR$\uparrow$ & $D_s$$\downarrow$ & $D_\lambda$$\downarrow$ & ERGAS$\downarrow$ & SCC$\uparrow$ & SAM$\downarrow$ & Q8$\uparrow$ & PSNR$\uparrow$ & SSIM$\uparrow$ \\
    \hline
    PanNet \cite{yang2017pannet} & 0.918 $\pm$ 0.031 & 0.049 $\pm$ 0.019 & 0.035 $\pm$ 0.014 & 2.538 $\pm$ 0.597 & 0.979 $\pm$ 0.006 & 3.402 $\pm$ 0.672 & 0.913 $\pm$ 0.087 & 36.148 $\pm$ 1.958 & 0.966 $\pm$ 0.011 \\
    MSDCNN \cite{yuan2018multiscale} & 0.924 $\pm$ 0.030 & 0.050 $\pm$ 0.020 & 0.028 $\pm$ 0.013 & 2.489 $\pm$ 0.620 & 0.979 $\pm$ 0.007 & 3.300 $\pm$ 0.654 & 0.914 $\pm$ 0.087 & 36.329 $\pm$ 1.748 & 0.967 $\pm$ 0.010 \\
    FusionNet \cite{wu2021dynamic} & 0.920 $\pm$ 0.030 & 0.053 $\pm$ 0.021 & 0.029 $\pm$ 0.011 & 2.428 $\pm$ 0.621 & 0.981 $\pm$ 0.007 & 3.188 $\pm$ 0.628 & 0.916 $\pm$ 0.087 & 36.569 $\pm$ 1.666 & 0.968 $\pm$ 0.009 \\
    LAGNet \cite{jin2022lagconv} & 0.915 $\pm$ 0.033 & 0.055 $\pm$ 0.023 & 0.033 $\pm$ 0.012 & 2.380 $\pm$ 0.617 & 0.981 $\pm$ 0.007 & 3.153 $\pm$ 0.608 & 0.916 $\pm$ 0.087 & 36.732 $\pm$ 1.723 & 0.970 $\pm$ 0.009 \\
    S2DBPN \cite{zhang2023spatial} & 0.946 $\pm$ 0.018 & 0.030 $\pm$ 0.010 & 0.025 $\pm$ 0.010 & 2.245 $\pm$ 0.541 & 0.985 $\pm$ 0.005 & 3.019 $\pm$ 0.588 & 0.917 $\pm$ 0.091 & 37.216 $\pm$ 1.888 & 0.972 $\pm$ 0.009 \\
    PanDiff \cite{meng2023pandiff} & 0.952 $\pm$ 0.009 & 0.034 $\pm$ 0.005 & {\color{red}{\textbf{0.014}}} $\pm$ 0.005 & 2.276 $\pm$ 0.545 & 0.984 $\pm$ 0.004 & 3.058 $\pm$ 0.567 & 0.913 $\pm$ 0.084 & 37.029 $\pm$ 1.796 & 0.971 $\pm$ 0.008 \\
    DCPNet \cite{zhang2024dcpnet} & 0.923 $\pm$ 0.027 & 0.036 $\pm$ 0.012 & 0.043 $\pm$ 0.018 & 2.301 $\pm$ 0.569 & 0.984 $\pm$ 0.005 & 3.083 $\pm$ 0.537 & 0.915 $\pm$ 0.092 & 37.009 $\pm$ 1.735 & 0.972 $\pm$ 0.008 \\
    TMDiff \cite{xing2024empower} & 0.924 $\pm$ 0.015 & 0.059 $\pm$ 0.009 & 0.018 $\pm$ 0.007 & 2.151 $\pm$ 0.458 & 0.986 $\pm$ 0.004 & 2.885 $\pm$ 0.549 & 0.915 $\pm$ 0.086 & 37.477 $\pm$ 1.923 & 0.973 $\pm$ 0.008 \\
    CANConv \cite{duan2024content} & 0.951 $\pm$ 0.013 & 0.030 $\pm$ 0.008 & 0.020 $\pm$ 0.008 & 2.163 $\pm$ 0.481 & 0.985 $\pm$ 0.005 & 2.927 $\pm$ 0.536 & 0.918 $\pm$ 0.082 & 37.441 $\pm$ 1.788 & 0.973 $\pm$ 0.008 \\
    \textbf{PAN-Crafter} & {\color{red}{\textbf{0.958}}} $\pm$ 0.009 & {\color{red}{\textbf{0.027}}} $\pm$ 0.004 & 0.016 $\pm$ 0.006 & {\color{red}{\textbf{2.040}}} $\pm$ 0.459 & {\color{red}{\textbf{0.988}}} $\pm$ 0.003 & {\color{red}{\textbf{2.787}}} $\pm$ 0.523 & {\color{red}{\textbf{0.922}}} $\pm$ 0.082 & {\color{red}{\textbf{37.956}}} $\pm$ 1.771 & {\color{red}{\textbf{0.976}}} $\pm$ 0.006 \\
    \Xhline{2\arrayrulewidth}
    \end{tabular}}
    \caption{Quantitative comparison of deep learning-based PS methods on the WV3 dataset. \textbf{\color{red}{Red}} indicates the best performance.}
    \label{tab:wv3_detail}
\end{table*}

\begin{table*}[tbp]
    \scriptsize
    \centering
    \resizebox{1.0\textwidth}{!}{ 
    \def\arraystretch{1.2}
    \begin{tabular}{l|ccc|cccccc}
    \Xhline{2\arrayrulewidth}
    \textbf{GF2} Dataset & \multicolumn{3}{c|}{Full-Resolution} & \multicolumn{6}{c}{Reduced-Resolution} \\
    \hline
    Methods & HQNR$\uparrow$ & $D_s$$\downarrow$ & $D_\lambda$$\downarrow$ & ERGAS$\downarrow$ & SCC$\uparrow$ & SAM$\downarrow$ & Q4$\uparrow$ & PSNR$\uparrow$ & SSIM$\uparrow$ \\
    \hline
    PanNet \cite{yang2017pannet} & 0.929 $\pm$ 0.013 & 0.052 $\pm$ 0.009 & 0.020 $\pm$ 0.012 & 1.038 $\pm$ 0.214 & 0.975 $\pm$ 0.006 & 1.050 $\pm$ 0.209 & 0.963 $\pm$ 0.009 & 39.197 $\pm$ 2.009 & 0.959 $\pm$ 0.011 \\
    MSDCNN \cite{yuan2018multiscale} & 0.898 $\pm$ 0.016 & 0.079 $\pm$ 0.011 & 0.026 $\pm$ 0.014 & 0.862 $\pm$ 0.141 & 0.983 $\pm$ 0.003 & 0.946 $\pm$ 0.166 & 0.972 $\pm$ 0.009 & 40.730 $\pm$ 1.564 & 0.971 $\pm$ 0.006 \\
    FusionNet\cite{wu2021dynamic} & 0.865 $\pm$ 0.018 & 0.105 $\pm$ 0.013 & 0.034 $\pm$ 0.013 & 0.960 $\pm$ 0.193 & 0.980 $\pm$ 0.005 & 0.971 $\pm$ 0.195 & 0.967 $\pm$ 0.008 & 39.866 $\pm$ 1.955 & 0.966 $\pm$ 0.009 \\
    LAGNet \cite{jin2022lagconv} & 0.895 $\pm$ 0.021 & 0.078 $\pm$ 0.013 & 0.030 $\pm$ 0.014 & 0.816 $\pm$ 0.121 & 0.985 $\pm$ 0.003 & 0.886 $\pm$ 0.140 & 0.974 $\pm$ 0.009 & 41.147 $\pm$ 1.384 & 0.974 $\pm$ 0.005 \\
    S2DBPN \cite{zhang2023spatial} & 0.935 $\pm$ 0.011 & 0.046 $\pm$ 0.007 & 0.020 $\pm$ 0.012 & 0.686 $\pm$ 0.125 & 0.990 $\pm$ 0.002 & 0.772 $\pm$ 0.149 & 0.981 $\pm$ 0.007 & 42.686 $\pm$ 1.676 & 0.980 $\pm$ 0.005 \\
    PanDiff \cite{meng2023pandiff} & 0.936 $\pm$ 0.011 & 0.045 $\pm$ 0.009 & 0.020 $\pm$ 0.014 & 0.674 $\pm$ 0.110 & 0.990 $\pm$ 0.002 & 0.767 $\pm$ 0.134 & 0.981 $\pm$ 0.007 & 42.827 $\pm$ 1.462 & 0.980 $\pm$ 0.005 \\
    DCPNet \cite{zhang2024dcpnet} & 0.953 $\pm$ 0.019 & 0.024 $\pm$ 0.008 & 0.024 $\pm$ 0.022 & 0.724 $\pm$ 0.138 & 0.988 $\pm$ 0.003 & 0.806 $\pm$ 0.153 & 0.980 $\pm$ 0.007 & 42.312 $\pm$ 1.682 & 0.979 $\pm$ 0.005 \\
    TMDiff \cite{xing2024empower} & 0.942 $\pm$ 0.016 & 0.030 $\pm$ 0.010 & 0.029 $\pm$ 0.011 & 0.754 $\pm$ 0.143 & 0.988 $\pm$ 0.003 & 0.764 $\pm$ 0.155 & 0.979 $\pm$ 0.007 & 41.896 $\pm$ 1.765 & 0.977 $\pm$ 0.005 \\
    CANConv \cite{duan2024content} & 0.919 $\pm$ 0.011 & 0.063 $\pm$ 0.009 & {\color{red}{\textbf{0.019}}} $\pm$ 0.010 & 0.653 $\pm$ 0.124 & 0.991 $\pm$ 0.002 & 0.722 $\pm$ 0.138 & 0.983 $\pm$ 0.006 & 43.166 $\pm$ 1.705 & 0.982 $\pm$ 0.004 \\
    \textbf{PAN-Crafter} & {\color{red}{\textbf{0.964}}} $\pm$ 0.015 & {\color{red}{\textbf{0.017}}} $\pm$ 0.007 & 0.020 $\pm$ 0.013 & {\color{red}{\textbf{0.552}}} $\pm$ 0.093 & {\color{red}{\textbf{0.994}}} $\pm$ 0.001 & {\color{red}{\textbf{0.596}}} $\pm$ 0.110 & {\color{red}{\textbf{0.988}}} $\pm$ 0.006 & {\color{red}{\textbf{45.076}}} $\pm$ 1.610 & {\color{red}{\textbf{0.988}}} $\pm$ 0.003 \\
    \Xhline{2\arrayrulewidth}
    \end{tabular}}
    \caption{Quantitative comparison of deep learning-based PS methods on the GF2 dataset. \textbf{\color{red}{Red}} indicates the best performance.}
    \label{tab:gf2_detail}
\end{table*}

\begin{table*}[tbp]
    \scriptsize
    \centering
    \resizebox{1.0\textwidth}{!}{ 
    \def\arraystretch{1.2}
    \begin{tabular}{l|ccc|cccccc}
    \Xhline{2\arrayrulewidth}
    \textbf{QB} Dataset & \multicolumn{3}{c|}{Full-Resolution} & \multicolumn{6}{c}{Reduced-Resolution} \\
    \hline
    Methods & HQNR$\uparrow$ & $D_s$$\downarrow$ & $D_\lambda$$\downarrow$ & ERGAS$\downarrow$ & SCC$\uparrow$ & SAM$\downarrow$ & Q4$\uparrow$ & PSNR$\uparrow$ & SSIM$\uparrow$ \\
    \hline
    PanNet \cite{yang2017pannet} & 0.851 $\pm$ 0.035 & 0.092 $\pm$ 0.021 & 0.063 $\pm$ 0.019 & 4.856 $\pm$ 0.590 & 0.966 $\pm$ 0.015 & 5.273 $\pm$ 0.946 & 0.911 $\pm$ 0.094 & 35.563 $\pm$ 1.930 & 0.939 $\pm$ 0.012 \\
    MSDCNN \cite{yuan2018multiscale} & 0.888 $\pm$ 0.037 & 0.058 $\pm$ 0.027 & 0.058 $\pm$ 0.014 & 4.074 $\pm$ 0.244 & 0.977 $\pm$ 0.010 & 4.828 $\pm$ 0.824 & 0.925 $\pm$ 0.098 & 37.040 $\pm$ 1.778 & 0.954 $\pm$ 0.007 \\
    FusionNet \cite{wu2021dynamic} & 0.853 $\pm$ 0.041 & 0.079 $\pm$ 0.025 & 0.074 $\pm$ 0.022 & 4.183 $\pm$ 0.266 & 0.975 $\pm$ 0.011 & 4.892 $\pm$ 0.822 & 0.923 $\pm$ 0.100 & 36.821 $\pm$ 1.765 & 0.952 $\pm$ 0.007 \\
    LAGNet \cite{jin2022lagconv} & 0.892 $\pm$ 0.024 & {\color{red}{\textbf{0.035}}} $\pm$ 0.009 & 0.075 $\pm$ 0.019 & 3.845 $\pm$ 0.323 & 0.980 $\pm$ 0.009 & 4.682 $\pm$ 0.785 & 0.930 $\pm$ 0.095 & 37.565 $\pm$ 1.721 & 0.958 $\pm$ 0.006 \\
    S2DBPN \cite{zhang2023spatial} & 0.908 $\pm$ 0.044 & 0.036 $\pm$ 0.023 & 0.059 $\pm$ 0.026 & 3.956 $\pm$ 0.291 & 0.980 $\pm$ 0.008 & 4.849 $\pm$ 0.822 & 0.928 $\pm$ 0.093 & 37.314 $\pm$ 1.782 & 0.956 $\pm$ 0.006 \\
    PanDiff \cite{meng2023pandiff} & 0.919 $\pm$ 0.010 & 0.055 $\pm$ 0.012 & {\color{red}{\textbf{0.028}}} $\pm$ 0.011 & 3.723 $\pm$ 0.280 & 0.982 $\pm$ 0.007 & 4.611 $\pm$ 0.768 & 0.935 $\pm$ 0.084 & 37.842 $\pm$ 1.721 & 0.959 $\pm$ 0.006 \\
    DCPNet \cite{zhang2024dcpnet} & 0.880 $\pm$ 0.013 & 0.073 $\pm$ 0.013 & 0.051 $\pm$ 0.017 & 3.618 $\pm$ 0.313 & 0.983 $\pm$ 0.010 & {\color{red}{\textbf{4.420}}} $\pm$ 0.710 & 0.935 $\pm$ 0.095 & 38.079 $\pm$ 1.454 & {\color{red}{\textbf{0.963}}} $\pm$ 0.004 \\
    TMDiff \cite{xing2024empower} & 0.901 $\pm$ 0.011 & 0.068 $\pm$ 0.012 & 0.034 $\pm$ 0.016 & 3.804 $\pm$ 0.279 & 0.981 $\pm$ 0.008 & 4.627 $\pm$ 0.814 & 0.930 $\pm$ 0.096 & 37.642 $\pm$ 1.831 & 0.958 $\pm$ 0.006 \\
    CANConv \cite{duan2024content} & 0.893 $\pm$ 0.010 & 0.070 $\pm$ 0.017 & 0.039 $\pm$ 0.012 & 3.740 $\pm$ 0.304 & 0.982 $\pm$ 0.007 & 4.554 $\pm$ 0.788 & 0.935 $\pm$ 0.087 & 37.795 $\pm$ 1.801 & 0.960 $\pm$ 0.006 \\
    \textbf{PAN-Crafter} & {\color{red}{\textbf{0.920}}} $\pm$ 0.027 & 0.039 $\pm$ 0.020 & 0.043 $\pm$ 0.011 & {\color{red}{\textbf{3.570}}} $\pm$ 0.286 & {\color{red}{\textbf{0.984}}} $\pm$ 0.008 & 4.426 $\pm$ 0.740 & {\color{red}{\textbf{0.938}}} $\pm$ 0.087 & {\color{red}{\textbf{38.195}}} $\pm$ 1.597 & {\color{red}{\textbf{0.963}}} $\pm$ 0.005 \\
    \Xhline{2\arrayrulewidth}
    \end{tabular}}
    \caption{Quantitative comparison of deep learning-based PS methods on the QB dataset. \textbf{\color{red}{Red}} indicates the best performance.}
    \label{tab:qb_detail}
\end{table*}

\begin{table*}[tbp]
    \scriptsize
    \centering
    \resizebox{1.0\textwidth}{!}{ 
    \def\arraystretch{1.2}
    \begin{tabular}{l|ccc|cccccc}
    \Xhline{2\arrayrulewidth}
    \textbf{WV2} Dataset & \multicolumn{3}{c|}{Full-Resolution (\textbf{Unseen} satellite dataset)} & \multicolumn{6}{c}{Reduced-Resolution (\textbf{Unseen} satellite dataset)} \\
    \hline
    Methods & HQNR$\uparrow$ & $D_s$$\downarrow$ & $D_\lambda$$\downarrow$ & ERGAS$\downarrow$ & SCC$\uparrow$ & SAM$\downarrow$ & Q8$\uparrow$ & PSNR$\uparrow$ & SSIM$\uparrow$ \\
    \hline
    PanNet \cite{yang2017pannet} & 0.875 $\pm$ 0.064 & 0.032 $\pm$ 0.005 & 0.096 $\pm$ 0.066 & 5.481 $\pm$ 0.326 & 0.876 $\pm$ 0.018 & 7.040 $\pm$ 0.417 & 0.786 $\pm$ 0.084 & 27.120 $\pm$ 1.827 & 0.770 $\pm$ 0.053 \\
    MSDCNN \cite{yuan2018multiscale} & 0.862 $\pm$ 0.050 & 0.029 $\pm$ 0.013 & 0.113 $\pm$ 0.041 & 4.930 $\pm$ 0.378 & 0.905 $\pm$ 0.009 & 5.898 $\pm$ 0.490 & 0.812 $\pm$ 0.090 & 27.901 $\pm$ 1.812 & 0.804 $\pm$ 0.040 \\
    FusionNet \cite{wu2021dynamic} & 0.862 $\pm$ 0.034 & 0.038 $\pm$ 0.005 & 0.104 $\pm$ 0.032 & 5.100 $\pm$ 0.367 & 0.902 $\pm$ 0.011 & 6.118 $\pm$ 0.533 & 0.786 $\pm$ 0.083 & 27.616 $\pm$ 1.765 & 0.788 $\pm$ 0.042 \\
    LAGNet \cite{jin2022lagconv} & 0.902 $\pm$ 0.045 & {\color{red}{\textbf{0.024}}} $\pm$ 0.018 & 0.076 $\pm$ 0.032 & 5.133 $\pm$ 0.432 & 0.885 $\pm$ 0.015 & 6.094 $\pm$ 0.559 & 0.792 $\pm$ 0.081 & 27.525 $\pm$ 2.008 & 0.777 $\pm$ 0.054 \\
    S2DBPN \cite{zhang2023spatial} & 0.813 $\pm$ 0.066 & 0.065 $\pm$ 0.019 & 0.129 $\pm$ 0.080 & 5.703 $\pm$ 0.257 & 0.915 $\pm$ 0.011 & 7.063 $\pm$ 0.421 & 0.805 $\pm$ 0.092 & 26.748 $\pm$ 1.892 & 0.804 $\pm$ 0.041 \\
    DCPNet \cite{zhang2024dcpnet} & 0.797 $\pm$ 0.134 & 0.034 $\pm$ 0.022 & 0.176 $\pm$ 0.129 & 5.507 $\pm$ 0.264 & {\color{red}{\textbf{0.931}}} $\pm$ 0.009 & 10.174 $\pm$ 1.115 & 0.843 $\pm$ 0.094 & 27.063 $\pm$ 1.541 & 0.855 $\pm$ 0.021 \\
    PanDiff \cite{meng2023pandiff} & 0.932 $\pm$ 0.019 & 0.043 $\pm$ 0.010 & 0.026 $\pm$ 0.019 & 4.291 $\pm$ 0.418 & 0.916 $\pm$ 0.010 & 5.430 $\pm$ 0.601 & 0.840 $\pm$ 0.087 & 28.964 $\pm$ 1.709 & 0.832 $\pm$ 0.033 \\
    TMDiff \cite{xing2024empower} & 0.874 $\pm$ 0.013 & 0.088 $\pm$ 0.021 & 0.042 $\pm$ 0.020 & 5.157 $\pm$ 0.604 & 0.875 $\pm$ 0.008 & 6.087 $\pm$ 0.786 & 0.777 $\pm$ 0.079 & 27.473 $\pm$ 1.634 & 0.762 $\pm$ 0.045 \\
    CANConv \cite{duan2024content} & 0.876 $\pm$ 0.044 & 0.060 $\pm$ 0.022 & 0.068 $\pm$ 0.049 & 4.328 $\pm$ 0.413 & 0.918 $\pm$ 0.008 & 5.481 $\pm$ 0.595 & 0.841 $\pm$ 0.087 & 29.005 $\pm$ 1.719 & 0.837 $\pm$ 0.031 \\
    \textbf{PAN-Crafter} & {\color{red}{\textbf{0.942}}} $\pm$ 0.019 & 0.036 $\pm$ 0.010 & {\color{red}{\textbf{0.022}}} $\pm$ 0.008 & {\color{red}{\textbf{4.169}}} $\pm$ 0.397 & 0.924 $\pm$ 0.009 & {\color{red}{\textbf{5.078}}} $\pm$ 0.561 & {\color{red}{\textbf{0.846}}} $\pm$ 0.085 & {\color{red}{\textbf{29.276}}} $\pm$ 1.621 & {\color{red}{\textbf{0.839}}} $\pm$ 0.029 \\
    \Xhline{2\arrayrulewidth}
    \end{tabular}}
    \caption{Quantitative comparison of deep learning-based PS methods on the unseen WV2 dataset. All models are trained on WV3 and evaluated on WV2 to assess real-world generalization. \textbf{\color{red}{Red}} indicate the best performance in each metric.}
    \label{tab:wv2_detail}
\end{table*}

\begin{table*}[tbp]
    \scriptsize
    \centering
    \resizebox{0.8\textwidth}{!}{
    \def\arraystretch{1.2}
    \begin{tabular}{l|cccccccc}
    \Xhline{2\arrayrulewidth}
    Methods & LAGConv \cite{jin2022lagconv} & S2DBPN \cite{zhang2023spatial} & PanDiff \cite{meng2023pandiff} & DCPNet \cite{zhang2024dcpnet} & TMDiff \cite{xing2024empower} & CANConv \cite{duan2024content} & \textbf{PAN-Crafter} \\
    \hline
    Time (s) & 0.004 & 0.005 & 2.955 & 0.109 & 9.997 & 0.451 & 0.009 \\
    Memory (MB) & 3360.1 & 2444.0 & 2383.6 & 7386.8 & 10147.4 & 2777.6 & 1751.9 \\
    FLOPs (G) & 8.43 & 158.94 & 62.07 & 105.40 & 1284.42 & 52.21 & 79.03 \\
    Params. (M) & 0.15 & 16.19 & 9.52 & 1.414 & 154.10 & 0.79 & 7.17 \\
    \Xhline{2\arrayrulewidth}
    \end{tabular}}
    \caption{Computational efficiency comparison of deep learning-based PS methods. We report inference time (s), memory usage (MB), FLOPs (G), and parameter count (M).}
    \label{tab:complex}
\end{table*}

\begin{table*}[tbp]
    \scriptsize
    \centering
    \resizebox{0.9\textwidth}{!}{
    \def\arraystretch{1.2}
    \begin{tabular}{c|c|ccc|cccccc|cc}
    \Xhline{2\arrayrulewidth}
    \multicolumn{2}{c|}{\textbf{WV3} Dataset} & \multicolumn{3}{c|}{Full-Resolution} & \multicolumn{6}{c|}{Reduced-Resolution} & \multirow{2}{*}{\makecell{Inference\\Time$\downarrow$ (s)}} & \multirow{2}{*}{\makecell{Memory$\downarrow$\\(GB)}} \\
    \cline{1-11}
    CM3A & MARs & HQNR$\uparrow$ & $D_{s}$$\downarrow$ & $D_{\lambda}$$\downarrow$ & ERGAS$\downarrow$ & SCC$\uparrow$ & SAM$\downarrow$ & Q8$\uparrow$ & PSNR$\uparrow$ & SSIM$\uparrow$ \\
    \hline
    & & 0.948 & 0.035 & 0.018 & 2.232 & 0.985 & 2.980 & 0.913 & 37.245 & 0.972 & 0.006 & 1.537 \\
    \checkmark & & 0.949 & 0.035 & 0.016 & 2.212 & 0.985 & 2.970 & 0.915 & 37.285 & 0.973 & 0.007 & 1.556 \\
    & \checkmark & 0.956 & 0.028 & 0.017 & 2.122 & 0.987 & 2.873 & 0.919 & 37.602 & 0.974 & 0.009 & 1.701 \\
    \checkmark & \checkmark & {\color{red}{\textbf{0.958}}} & {\color{red}{\textbf{0.027}}} & {\color{red}{\textbf{0.016}}} &{\color{red}{\textbf{2.040}}} & {\color{red}{\textbf{0.988}}} & {\color{red}{\textbf{2.787}}} & {\color{red}{\textbf{0.922}}} & {\color{red}{\textbf{37.956}}} & {\color{red}{\textbf{0.976}}} & 0.009 & 1.711 \\
    \Xhline{2\arrayrulewidth}
    \end{tabular}}
    \caption{Ablation studies on CM3A and MARs on the WV3 dataset.}
  \label{tab:suppe_ablation_wv3}
\end{table*}

\begin{table*}[tbp]
    \scriptsize
    \centering
    \resizebox{0.7\textwidth}{!}{
    \def\arraystretch{1.2}
    \begin{tabular}{c|c|ccc|cccccc}
    \Xhline{2\arrayrulewidth}
    \multicolumn{2}{c|}{\textbf{GF2} Dataset} & \multicolumn{3}{c|}{Full-Resolution} & \multicolumn{6}{c}{Reduced-Resolution} \\
    \cline{1-11}
    CM3A & MARs & HQNR$\uparrow$ & $D_{s}$$\downarrow$ & $D_{\lambda}$$\downarrow$ & ERGAS$\downarrow$ & SCC$\uparrow$ & SAM$\downarrow$ & Q4$\uparrow$ & PSNR$\uparrow$ & SSIM$\uparrow$ \\
    \hline
    & & 0.959 & 0.021 & 0.021 & 0.632 & 0.992 & 0.723 & 0.984 & 43.476 & 0.984 \\
    \checkmark & & 0.953 & 0.025 & 0.023 & 0.624 & 0.992 & 0.718 & 0.984 & 43.618 & 0.984 \\
    & \checkmark & 0.945 & 0.032 & 0.023 & 0.574 & 0.993 & 0.651 & 0.986 & 44.298 & 0.986 \\
    \checkmark & \checkmark & {\color{red}{\textbf{0.964}}} & {\color{red}{\textbf{0.017}}} & {\color{red}{\textbf{0.020}}} &{\color{red}{\textbf{0.552}}} & {\color{red}{\textbf{0.994}}} & {\color{red}{\textbf{0.596}}} & {\color{red}{\textbf{0.988}}} & {\color{red}{\textbf{45.076}}} & {\color{red}{\textbf{0.988}}} \\
    \Xhline{2\arrayrulewidth}
    \end{tabular}}
    \caption{Ablation studies on CM3A and MARs on the GF2 dataset.}
  \label{tab:suppe_ablation_gf2}
\end{table*}

\begin{table*}[tbp]
    \scriptsize
    \centering
    \resizebox{0.7\textwidth}{!}{
    \def\arraystretch{1.2}
    \begin{tabular}{c|c|ccc|cccccc}
    \Xhline{2\arrayrulewidth}
    \multicolumn{2}{c|}{\textbf{QB} Dataset} & \multicolumn{3}{c|}{Full-Resolution} & \multicolumn{6}{c}{Reduced-Resolution} \\
    \cline{1-11}
    CM3A & MARs & HQNR$\uparrow$ & $D_{s}$$\downarrow$ & $D_{\lambda}$$\downarrow$ & ERGAS$\downarrow$ & SCC$\uparrow$ & SAM$\downarrow$ & Q4$\uparrow$ & PSNR$\uparrow$ & SSIM$\uparrow$ \\
    \hline
    & & 0.856 & 0.086 & 0.064 & 4.907 & 0.977 & 5.200 & 0.923 & 35.476 & 0.947 \\
    \checkmark & & 0.879 & 0.062 & 0.063 & 4.869 & 0.975 & 5.168 & 0.922 & 35.538 & 0.947 \\
    & \checkmark & 0.896 & 0.047 & 0.060 & 3.857 & 0.980 & 4.661 & 0.930 & 37.557 & 0.959 \\
    \checkmark & \checkmark & {\color{red}{\textbf{0.920}}} & {\color{red}{\textbf{0.039}}} & {\color{red}{\textbf{0.043}}} &{\color{red}{\textbf{3.570}}} & {\color{red}{\textbf{0.984}}} & {\color{red}{\textbf{4.426}}} & {\color{red}{\textbf{0.938}}} & {\color{red}{\textbf{38.195}}} & {\color{red}{\textbf{0.963}}} \\
    \Xhline{2\arrayrulewidth}
    \end{tabular}}
    \caption{Ablation studies on CM3A and MARs on the QB dataset.}
  \label{tab:suppe_ablation_qb}
\end{table*}

\begin{table*}[tbp]
    \scriptsize
    \centering
    \resizebox{0.75\textwidth}{!}{
    \def\arraystretch{1.2}
    \begin{tabular} {c|c|c|ccc}
        \Xhline{2\arrayrulewidth}
        \multirow{2}{*}{$\bm{\alpha}$} & \multirow{2}{*}{$\bm{\beta}, \bm{\gamma}$} &  \multicolumn{4}{c}{\textbf{WV3} / \textbf{GF2} / \textbf{QB} Datasets} \\
        \cline{3-6}
        & & HQNR$\uparrow$ & ERGAS$\downarrow$ & SAM$\downarrow$ & PSNR$\uparrow$ \\
        \hline
        & & 0.945 / 0.951 / 0.908 & 2.214 / 0.623 / 3.758 & 2.901 / 0.642 / 4.523 & 37.210 / 44.321 / 37.842 \\
        \checkmark & & 0.949 / 0.957 / 0.915 & 2.150 / 0.589 / 3.669 & 2.829 / 0.618 / 4.472 & 37.562 / 44.758 / 38.021 \\
        & \checkmark & 0.947 / 0.956 / 0.913 & 2.185 / 0.601 / 3.690 & 2.841 / 0.624 / 4.488 & 37.433 / 44.612 / 37.935 \\
        \checkmark & \checkmark & {\color{red}{\textbf{0.958}}} / {\color{red}{\textbf{0.964}}} / {\color{red}{\textbf{0.920}}} & {\color{red}{\textbf{2.040}}} / {\color{red}{\textbf{0.552}}} / {\color{red}{\textbf{3.570}}} & {\color{red}{\textbf{2.787}}} / {\color{red}{\textbf{0.596}}} / {\color{red}{\textbf{4.426}}} & {\color{red}{\textbf{37.956}}} / {\color{red}{\textbf{45.076}}} / {\color{red}{\textbf{38.195}}} \\
        \Xhline{2\arrayrulewidth}
    \end{tabular}}
    \caption{Ablation studies on $\bm{\alpha}$, $\bm{\beta}$, and $\bm{\gamma}$ on the WV3, GF2, and QB datasets.}
    \label{tab:suppe_ablation_1}
\end{table*}

\begin{table*}[tbp]
    \scriptsize
    \centering
    \resizebox{0.8\textwidth}{!}{
    \def\arraystretch{1.2}
    \begin{tabular} {c|c|ccc|cc}
    \Xhline{2\arrayrulewidth}
    \multirow{2}{*}{$k$} & \multicolumn{6}{c}{\textbf{WV3} / \textbf{GF2} / \textbf{QB} Datasets} \\
    \cline{2-7}
    & HQNR$\uparrow$ & ERGAS$\downarrow$ & SAM$\downarrow$ & PSNR$\uparrow$ & Time$\downarrow$ & Memory$\downarrow$ \\
    \hline
    3 & {\color{red}{\textbf{0.958}}} / 0.964 / 0.920 & 2.040 / {\color{red}{\textbf{0.552}}} / {\color{red}{\textbf{3.570}}} & 2.787 / {\color{red}{\textbf{0.596}}} / {\color{red}{\textbf{4.426}}} & 37.956 / {\color{red}{\textbf{45.076}}} / 38.195 & {\color{red}{\textbf{0.009}}} & {\color{red}{\textbf{1.711}}} \\
    5 & 0.953 / {\color{red}{\textbf{0.965}}} / 0.919 & {\color{red}{\textbf{2.021}}} / 0.555 / 3.577 & {\color{red}{\textbf{2.785}}} / 0.600 / 4.433 & {\color{red}{\textbf{37.966}}} / 45.001 / {\color{red}{\textbf{38.201}}} & 0.019 & 3.429 \\
    7 & 0.955 / 0.961 / {\color{red}{\textbf{0.921}}} & 2.033 / 0.553 / 3.575 & 2.790 / 0.599 / 4.429 & 37.949 / 45.010 / 38.190 & 0.042 & 5.243 \\
    \Xhline{2\arrayrulewidth}
    \end{tabular}}
    \caption{Ablation studies on $k$ on the WV3, GF2, and QB datasets.}
    \label{tab:suppe_ablation_2}
\end{table*}

\begin{table*}[tbp]
    \scriptsize
    \centering
    \resizebox{0.7\textwidth}{!}{
    \def\arraystretch{1.2}
    \begin{tabular} {l|c|ccc}
        \Xhline{2\arrayrulewidth}
        \multirow{2}{*}{\makecell{Training\\Strategy}} & \multicolumn{4}{c}{\textbf{WV3} / \textbf{GF2} / \textbf{QB} Datasets} \\
        \cline{2-5}
        & HQNR$\uparrow$ & ERGAS$\downarrow$ & SAM$\downarrow$ & PSNR$\uparrow$ \\
        \hline
        w/o MARs & 0.949 / 0.953 / 0.879 & 2.212 / 0.624 / 4.869 & 2.970 / 0.718 / 5.168 & 37.285 / 43.618 / 35.538 \\
        Two-stage & 0.945 / 0.953 / 0.890 & 2.199 / 0.602 / 4.551 & 2.899 / 0.688 / 4.907 & 37.345 / 43.921 / 36.081 \\
        w/ MARs & {\color{red}{\textbf{0.958}}} / {\color{red}{\textbf{0.964}}} / {\color{red}{\textbf{0.920}}} & {\color{red}{\textbf{2.040}}} / {\color{red}{\textbf{0.552}}} / {\color{red}{\textbf{3.570}}} & {\color{red}{\textbf{2.787}}} / {\color{red}{\textbf{0.596}}} / {\color{red}{\textbf{4.426}}} & {\color{red}{\textbf{37.956}}} / {\color{red}{\textbf{45.076}}} / {\color{red}{\textbf{38.195}}} \\
        \Xhline{2\arrayrulewidth}
    \end{tabular}}
    \caption{Ablation studies on training strategy on the WV3, GF2, and QB datasets.}
    \label{tab:suppe_ablation_3}
\end{table*}

\begin{table*}[tbp]
    \scriptsize
    \centering
    \resizebox{0.8\textwidth}{!}{
    \def\arraystretch{1.2}
    \begin{tabular} {c|c|ccc|cc}
        \Xhline{2\arrayrulewidth}
        \multirow{2}{*}{\makecell{Layer\\Type}} & \multicolumn{6}{c}{\textbf{WV3} / \textbf{GF2} / \textbf{QB} Datasets} \\
        \cline{2-7}
        & HQNR$\uparrow$ & ERGAS$\downarrow$ & SAM$\downarrow$ & PSNR$\uparrow$ & Time$\downarrow$ & Memory$\downarrow$ \\
        \hline
        Attn. & {\color{red}{\textbf{0.948}}} / {\color{red}{\textbf{0.959}}} / {\color{red}{\textbf{0.856}}} & {\color{red}{\textbf{2.232}}} / {\color{red}{\textbf{0.632}}} / {\color{red}{\textbf{4.907}}} & {\color{red}{\textbf{2.980}}} / {\color{red}{\textbf{0.723}}} / {\color{red}{\textbf{5.200}}} & {\color{red}{\textbf{37.245}}} / {\color{red}{\textbf{43.476}}} / {\color{red}{\textbf{35.476}}} & 0.006 & 1.537 \\
        Conv. & 0.937 / 0.943 / 0.850 & 2.322 / 0.741 / 5.142 & 3.120 / 0.831 / 5.463 & 36.988 / 42.590 / 35.218 & {\color{red}{\textbf{0.004}}} & {\color{red}{\textbf{1.209}}} \\
        \Xhline{2\arrayrulewidth}
    \end{tabular}}
    \caption{Ablation studies on layer type on the WV3, GF2, and QB datasets.}
    \label{tab:suppe_ablation_4}
\end{table*}

\clearpage

{
    \small
    \bibliographystyle{ieeenat_fullname}
    \bibliography{main}
}

\end{document}